\documentclass{article}
\usepackage{frExamplee}
\usepackage{graphicx}
\usepackage{apalike}
\usepackage{setspace}
\usepackage{xcolor}
\usepackage{cleveref}

\usepackage{float}
\usepackage{subcaption}
\usepackage{multicol}
\usepackage{multirow}
\usepackage{array}
\usepackage{caption}
\usepackage{float}
\usepackage{multicol}
\usepackage{soul}
\usepackage{xfrac}
\usepackage{url}
\usepackage{afterpage}

\newcommand{\RomanNumeralCaps}[1]{\MakeUppercase{\romannumeral #1}}

\title{Congestion Analysis for the DARPA OFFSET CCAST Swarm }

\author{
Robert Brown \\
Collaborative Robotics and Intelligent \\
Systems Institute \\
Oregon State University \\
Corvallis OR 97331, USA \\
 \texttt{brownro3@oregonstate.edu} \\
\And
Julie A. Adams \\
Collaborative Robotics and Intelligent \\
Systems Institute \\
Oregon State University \\
Corvallis OR 97331, USA \\
 \texttt{julie.a.adams@oregonstate.edu}
}
\begin{document}

\maketitle

\begin{abstract}

The Defense Advanced Research Projects Agency's (DARPA) OFFensive Swam-Enabled Tactics program's goal of launching 250 unmanned aerial and ground vehicles from a limited sized launch zone was a daunting challenge. The swarm's aerial vehicles were primarily multi-rotor platforms, which can efficiently be launched en mass. Each field exercise expected the deployment of an even larger swarm. While the launch zone's spatial area increased with each field exercise, the relative space for each vehicle was not necessarily increased considering the increasing size of the swarm and the vehicles' associated GPS error. However, safe mission deployment and execution were expected. At the same time, achieving the mission goals required maximizing the efficiency of the swarm's performance, by reducing congestion that blocked vehicles from completing tactic assignments. Congestion analysis conducted before the final field exercise focused on adjusting various constraints to optimize the swarm's deployment without reducing safety. During the field exercise, data was collected that permitted analyzing the number and durations of individual vehicle blockages' impact on the resulting congestion. After the field exercise, additional analyses used the mission plan to validate the use of simulation for analyzing congestion.  
\end{abstract}

\section{Introduction}

The Defense Advanced Research Projects Agency (DARPA) OFFensive Swam-Enabled Tactics (OFFSET) program was designed to enable a very large heterogeneous swarm of unmanned air and ground vehicles in complex urban environments \cite{darpa_offset}. As swarm size increased, DARPA intentionally limited the launch zone size and allotted deployment time in order to ``encourage'' the teams to address swarm deployment logistics challenges. The OFFSET program's Command and Control of Aggregate Swarm Tactics (CCAST) team's swarm architecture was designed to enable a single operator to deploy and monitor a swarm of up to 250 unmanned vehicles for diverse missions \cite{ccast}. 

Over the course of the OFFSET program, the swarm size increased as the field exercises occurred at differing Department of Defense Combined Armed Collective Training Facilities (CACTF). Each CACTF presented different challenges when deploying a hardware swarm composed of heterogeneous ground and multi-rotor aerial vehicles. The CACTF's size and shape as well as its structures (e.g., buildings, light poles, power lines, street signs, and curbs), the designated launch/landing zone size, along with the swarm's size and composition influenced the distribution of vehicles and increased the likelihood of launch, en-route, and landing conflicts amongst the vehicles, in other words, \textit{congestion}. The challenge was determining how to deploy the swarm effectively while minimizing the congestion and navigation path planning conflicts. More specifically, these conflicts occurred when a large number of vehicles that rely on GPS localization, with a large localization error, deploy and return to a small area, called the \textit{launch zone}. This congestion can negatively impact the swarm's performance, delaying or interrupting mission plans as well as causing vehicles, particularly aerial vehicles, to deplete their batteries and shorten their deployable mission time. 

The CCAST team intended to deliver a fleet of at least a 250 hardware swarm for the final exercise; however, due to uncontrollable circumstances, CCAST's entire fleet consisted of 183 vehicles: 44 ground robots (UGVs), and 139 multi-rotor aerial vehicles (UAVs). The heterogeneous swarm consisted of relatively small, low cost (e.g., the most expensive being \$3,900) commercially available platforms, see Table \ref{Tab:Robots}, some of which  were augmented with necessary sensors and processing capabilities. While very capable, these vehicles have limitations compared to larger, more expensive robots, but the trade-off was to use inexpensive platforms in order to scale the swarm's size. 

\begin{table}[t]
\caption{CCAST's Robot Platforms, including cost, size, and number of each.} 
\label{Tab:Robots}
\center
\begin{tabular}{|m{1.13in}|m{1.13in}|m{1.13in}|m{1.13in}|m{1.13in}|}
  \hline
\center{\textbf{Aion Robotics R1}}  & \center{\textbf{3DR Solo}} & \center{\textbf{Uvify Ifo-S}} & \center{\textbf{Modal AI VOXL M500}} &  \textbf{Modal AI Micro-Seeker}\\
 \center{ \includegraphics[width=1.13in]{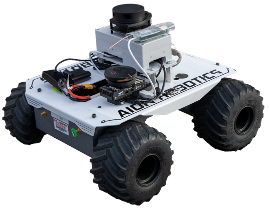}}  & \center{\includegraphics[width=1.13in]{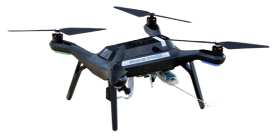}} & \center{\includegraphics[width=1.13in]{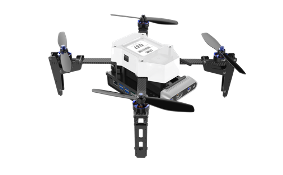}} & \center{\includegraphics[width=1.13in]{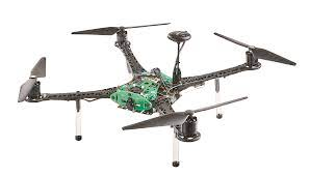}} &  
 \includegraphics[width=1.15in]{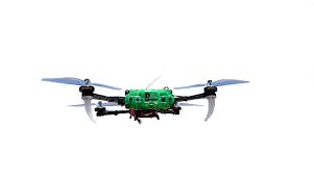} \\
  \hline\hline
\center{ \textasciitilde\$3,600 }&   
\center{ \textasciitilde\$750} &   
\center{ \textasciitilde\$3,900} &    
\center{ \textasciitilde\$2,300} &   \textasciitilde\$2700 \\ \hline
42.6cm x 48.6cm &   
25.9cm x 18.8cm &   
27.5cm x 27.5cm &
39.37cm x 39.37cm &    
19.1cm x 19.1cm \\ \hline
44 & 40 & 21 & 69 & 9 \\
\hline 
\end{tabular}
\end{table}

DARPA intentionally limited the launch zone size, challenging the ability to deploy all vehicles simultaneously. While the UGVs can detect and avoid UAVs positioned in the launch zone, it is not desirable for the UGVs to navigate through the UAVs. All vehicles self-localize via GPS, but the vehicles' smaller size, relative to up to a 5 meter (m) GPS error, resulted in an operational procedure to maintain a 5m distance between all vehicles within the launch zone. Further, vehicles may be blocked, or unable to plan a traversable navigation path, by other vehicles, either on the ground or in the air. Blocked UAVs are required to hover while replanning, which consumes more power and reduces their deployable time. Finally, the CACTF's built environment creates obstacles and choke points that can introduce vehicle congestion. These blockages, or congestion, can occur either en route, near task goals (e.g., approaching a building to surveil it), or over the launch zone when taking off or returning to launch (RTL). 
Given these constraints, the objective is to determine how to optimize deploying larger heterogeneous swarms within the constrained launch area in order  to achieve the field exercise's mission priorities. Achieving this objective requires investigating launch zone vehicle configurations, safety protocol variations (e.g., reducing the ``safe'' distance between platforms), and mission plan modifications.

\begin{figure}[htb]
\centering
\begin{subfigure}{.49\textwidth}
  \centering
  \includegraphics[width=\linewidth]{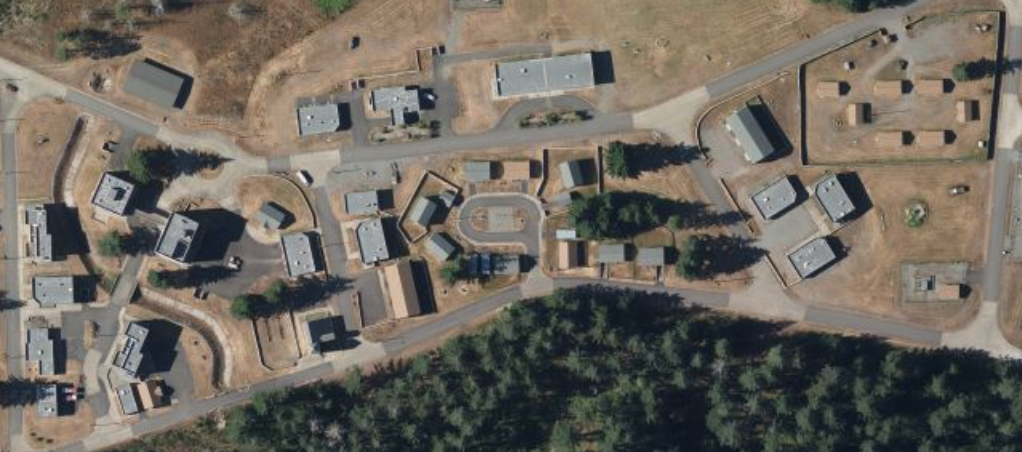}
    \caption{Joint Base Lewis McChord CACTF.}
  \label{fig:FX4CACTF}
\end{subfigure}%
\hfill
\begin{subfigure}{.3\textwidth}
  \centering
  \includegraphics[width=\linewidth]{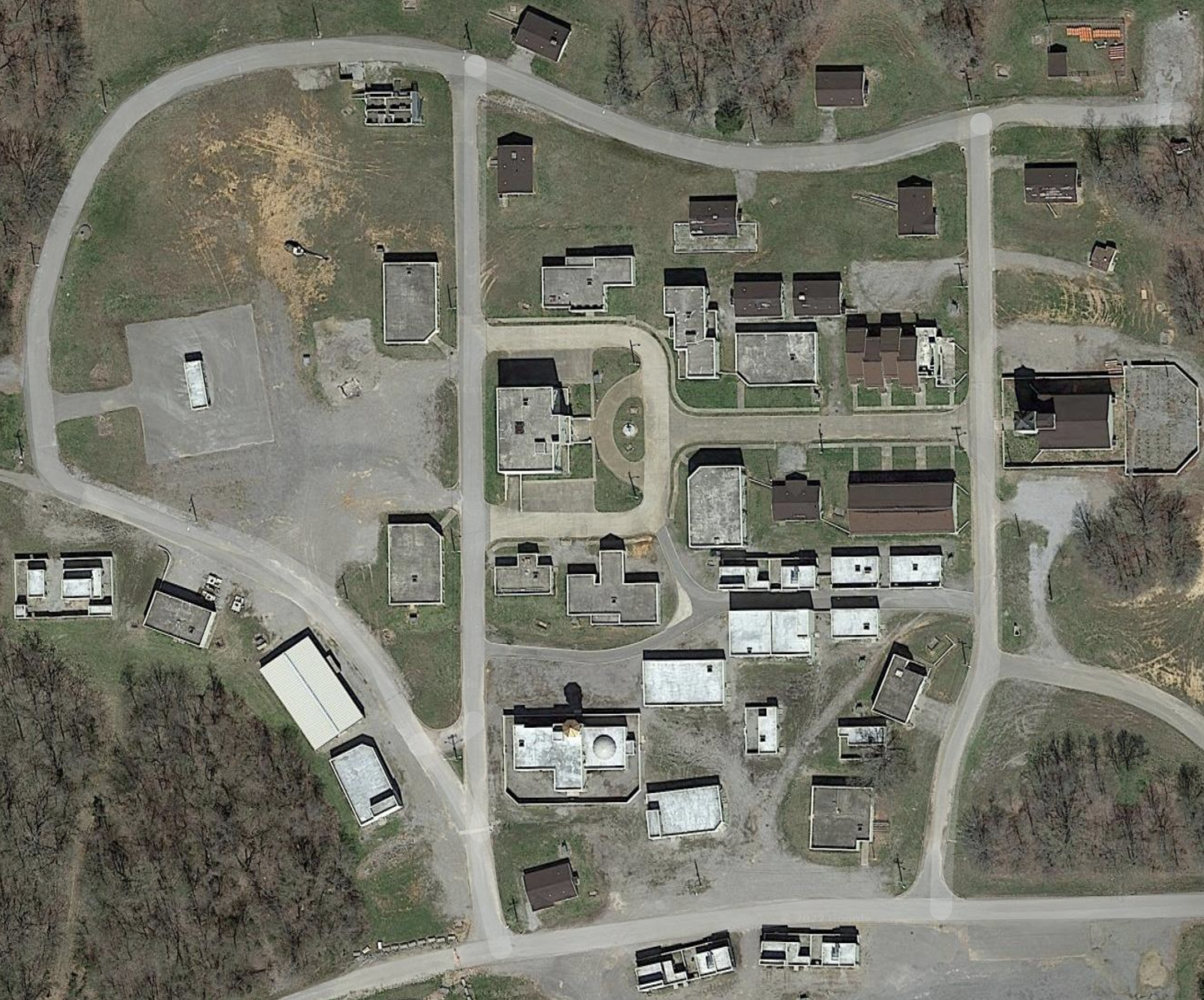}
    \caption{Fort Campbell CACTF.}
  \label{fig:FX6CACTF}
\end{subfigure}%

\caption{The field exercise CACTFs, not to scale.}
 \label{fig:CACTFs}
\end{figure}

Two CACTFs were analyzed. Joint Base Lewis-McChord's Leschi Town, shown in Figure \ref{fig:FX4CACTF}, the location of Field Exercise (FX) 4 was initially considered for FX6. Ultimately, Fort Campbell's Cassidy CACTF\footnote{Note, Field Exercise 5 was canceled.}, shown in Figure \ref{fig:FX6CACTF}, was chosen as the FX6 site. This manuscript presents analyses of actual and simulated missions, with the simulated mission results serving as a baseline for the FX6's actual results. Potential congestion has a more significant impact on UAVs' flight times and their contributions to achieving the mission scenario objectives; thus, the CCAST swarm's UAVs are the primary focus of this analysis. 

An overview of the CCAST swarm system is provided, followed by a review of the relevant congestion mitigation literature. The congestion analysis results for both CACTFs prior to FX6 are provided. A follow-up analysis, after FX6, investigates congestion that occurred during the FX6 to that derived from simulation trials using the same mission plans. CCAST's multi-fidelity swarm simulator was used to generate the results. The experimental methodology for each analysis is provided, along with the corresponding results. Discussions provide insights into the impact of congestion on mission progress and mitigation approaches. 

\section{CCAST Swarm System Architecture} \label{sec:system_architecture}

The CCAST swarm architecture has four primary components: the vehicles, a mission planner, the swarm commander interface (I3), and the swarm dispatcher \cite{ccast}. The mission planner is used prior to mission deployment and permits composing tactics into a mission plan. The tactics may require vehicles with specific capabilities or payloads and can incorporate inter-tactic ordering dependencies. 

The heterogeneous CCAST swarm consists of varying numbers of commercially available vehicles (i.e., 3DR Solos~\cite{3dr_solo}, Uvify Ifo-Ss~\cite{uvify_ifo}, Modal AI VOXL m500s~\cite{voxl_m500}, Modal AI Micro-Seekers~\cite{micro_seeker}, and Aion R1/R6 UGVs~\cite{aion_r1}) with varying sensing and computational processing capabilities, as detailed in Table \ref{Tab:Robots}. The CCAST architecture knows each vehicle's sensor and processing capabilities (e.g., Uvify Ifo-S's payloads permit indoor flight, 3DR Solos's payloads do not). The exact mission deployment hardware vehicle distribution depends on various factors, including each field exercise's available vehicles (e.g., the AI Modal UAVs were not part of the CCAST swarm at FX4), the mission plan, environmental conditions, etc. The FX6 CCAST swarm hardware composition is provided in the table; however, deployed swarm compositions varied during the FX6 shifts.  

Communication between I3, the swarm dispatcher, the vehicles, and I3 occurs over an LTE network, using a publish/subscribe protocol. Each vehicle's LTE modem allows it to communicate with the LTE basestation. The vehicles communicate telemetry data to the dispatcher, which relays that information to other vehicles and I3. This communication architecture relies on the vehicles' having direct line-of-sight with the LTE basestation in order to communicate data packets. The nature of the FXs' built CACTF environment necessitates the CCAST system's ability to be resilient to vehicles being unable to communicate with the rest of the system. This situation can occur as vehicles move throughout the dense urban environment, and buildings or trees block a vehicle's line-of-site to the LTE basestation. 

Prior to mission deployments, the CCAST Swarm Tactic Operations Mission Planner is used to prepare a relevant mission plan that seeks to achieve the mission objectives. The resulting plan can be evaluated and refined using virtual vehicles available in CCAST's multi-resolution swarm simulation. After the vehicles are staged in the FX's launch zone and powered on, the mission plan is instantiated, binding available vehicles on the LTE network to mission relevant roles or groups. When the mission plan tactics are instantiated, they are assigned to the appropriate vehicles that are spatially closest to the tactics goal location. 

The resulting mission plan is composed of relevant tactics from the CCAST Swarm Tactics Exchange extensible library. The mission plan may include phases that group tactics in order to achieve important mission goals (e.g., Phase I: information, surveillance, and reconnaissance, Phase II: Act on gathered intelligence to locate a verified hostile, Phase III: neutralize the verified hostile). This library incorporates tactics for surveilling structures or areas of interest, flocking, agent following, exploring building interiors, etc. The swarm vehicles are assigned tactics either as individuals or as a team. The vehicles can automatically swap in order to continue tactics when vehicle battery levels become too low \cite{DiehlDARS2022}. Once a tactic is assigned, the vehicles conduct on-board real-time navigation planning using extensions to the real-time, rapidly exploring random tree star (RT-RRT*) algorithm \cite{rt_rrt}. The RT-RRT* algorithm incorporates randomness when searching for potential paths, resulting in vehicles identifying different paths to achieve the swarm's mission objectives. 

The CCAST multi-resolution swarm simulator extends Microsoft Research's AirSim \cite{airsim} and facilitates rapid system development, pre-FX (e.g., congestion testing), and pre-mission (e.g., mission planning) analysis. The CCAST extensions permit both larger swarm scales, and simultaneous live/virtual vehicle deployments.  This simulator was leveraged to generate the reported congestion evaluation results. 

The CCAST team is assigned designated shifts for the FX swarm deployments. Early FX shifts are dedicated to shorter (i.e., 1.5 - 2 hours) system integration and dry runs, while later longer shifts (i.e., 2 - 3.5 hours) focus on ``playing'' the mission scenario. Once the CCAST hardware vehicles are positioned in the launch zone, the remaining system components are activated, the pre-mission brief has been conducted, and the vehicles are powered on, a shift's mission deployment can begin. 

The CCAST swarm is deployed and managed by a single human, the swarm commander, via a 3-D virtual reality-based interface (I3). At shift start, the swarm commander loads the mission plan and either executes the entire mission plan, or portions of (i.e., signals within) a multi-phase mission plan. The swarm commander can also generate tactic assignments that are explicitly or implicitly assigned to vehicles. The mission plan components and the swarm commander's generated tactics are communicated to the swarm dispatcher, which takes the necessary actions to communicate the tactics to the relevant vehicles. The swarm dispatcher coordinates inter-vehicle communication and relays vehicle telemetry to I3. 

If the swarm commander has not explicitly identified the vehicles to execute a tactic, the dispatcher automatically selects them from the available unassigned vehicles with the necessary capabilities that are spatially located closest to the tactic's goal location (e.g., the building to be surveilled). The allocated vehicles individually plan navigation paths and, when found, execute those paths. This navigation planning can fail for multiple reasons, such as a vehicle's path being blocked by another vehicle or the designated target position being unreachable. Using a CCAST generated 3D terrain elevation model that includes known structures and obstacles, the dispatcher is expected only to assign tactic goal execution points the vehicles can reach; however, congestion will occur when a vehicle is unable to plan a navigation path due to being blocked by one or more vehicles, structures, or obstacles.

\begin{figure}[htb]
\centering
  \includegraphics[width=0.49\linewidth]{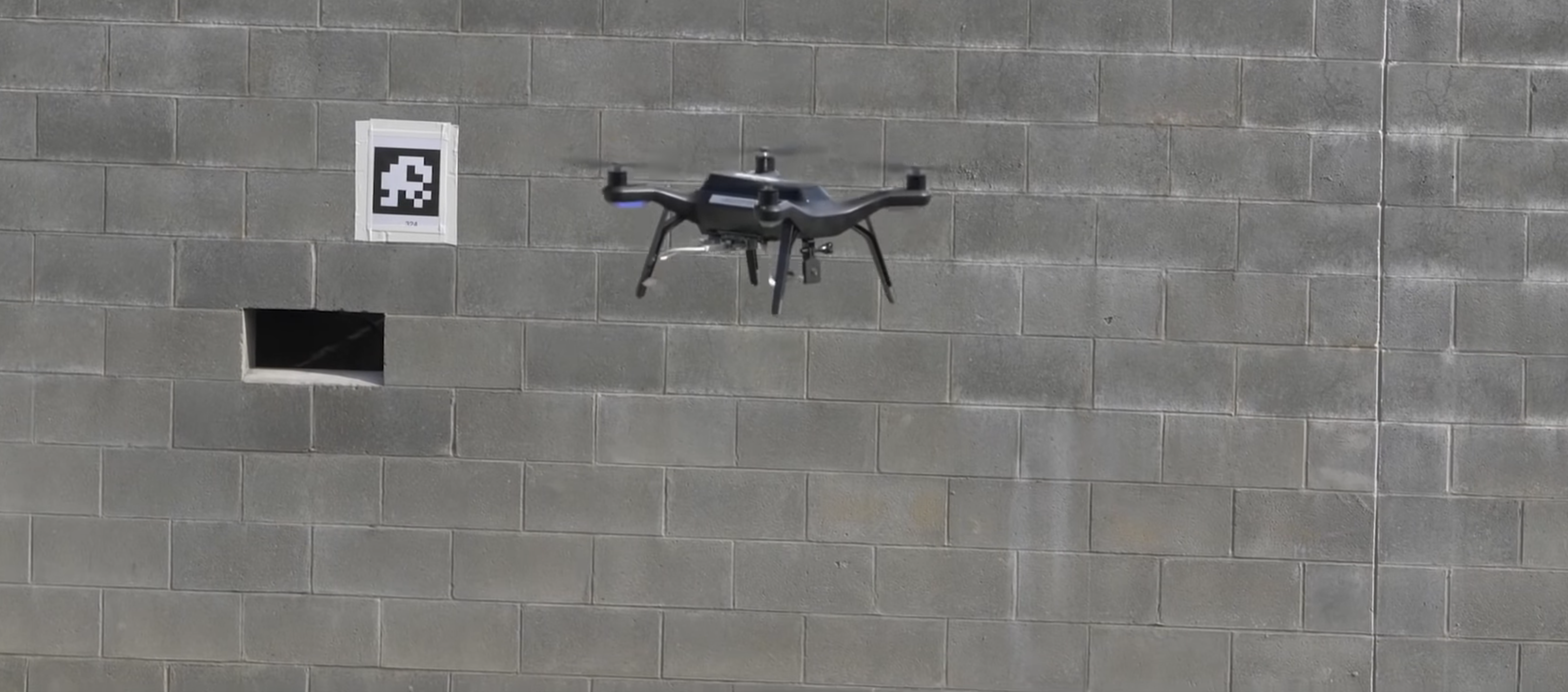}
\caption{A Building surveillance task, during which a UAV detects the artifact. Photo courtesy of DARPA.}
 \label{fig:SoloTag}
\end{figure}

While the CCAST swarm is capable of a wide variety of offensive, defensive, and surveillance tactics, the primary focus for the congestion analysis is a prevalent mission tactic, the Building surveillance tactic. The Building surveillance (surveil) tactic is central to gathering mission relevant intelligence. This tactic requires four UAVs with forward-facing cameras to investigate the sides of a building, and a fifth UAV with a downward-facing camera to investigate the same building's roof. The Phase I mission plan often includes large numbers of Building surveil tactics that are ``fired off'' simultaneously at shift start, which can generate a lot of vehicle movement, and resulting congestion. The Building surveil tactic execution begins with UAVs launching and ascending to a randomly assigned altitude, which was between 25m and 50m above ground level (AGL) during FX6. Simultaneously with these actions, each UAV begins planning a navigation path, and once a deconflicted navigable path is identified, the UAV beings moving toward the assigned building. While en route, the UAVs typically fly at an altitude safely above buildings and treetops, but during the surveillance execution, the UAVs descend to complete the task, as shown for one UAV in Figure~\ref{fig:SoloTag}. DARPA places April Tags \cite{april_tags}, representing different mission relevant artifacts (e.g., building ingress points, improvised explosive devices, locations of mission relevant information or high value targets) on horizontal and vertical surfaces throughout the CACTF, including inside and outside buildings. The Building surveil tactics' UAVs with forward-facing cameras must descend into the built environment, often within a few meters of the ground (e.g., 5m AGL), in order to sense any available mission artifacts located on the side of the building, as shown in Figure~\ref{fig:SoloTag}. Successful UAVs complete their Building surveil tactic and automatically return to the location from which they launched in the designated launch zone. However, many hazards (e.g., improvised explosive devices ) and adversarial (e.g., verified hostiles) artifacts exist that can neutralize the CCAST swarm vehicles, rendering them unable to continue executing the assigned tactic. Neutralized UAVs automatically return to the launch zone and land, later to be revived by a mobile medic. Further, the vehicles' consumption of the available battery power will shorten a UAV's deployment time. 

Vehicles can be assigned to complete multiple consecutive tactics, such as multiple Building surveils in a particular region of the CACTF, rather than automatically returning to the launch zone upon completion of a single tactic. Using this approach optimizes the speed of gathering intelligence, can potentially reduce congestion, and permits optimizing the usage of the UAVs' power source, while working towards achieving the mission objectives. A key factor is managing power usage. While not an issue for UGVs, whose batteries provided sufficient power for even the longest FX shifts, UAVs' batteries only support 10-20 minute flights. As a safety precaution, the UAVs are programmed to automatically RTL when their available battery level is reduced to the Battery RTL level. Tactics that require the UAVs to hover, such as Building surveils, consume more power than en-route flight maneuvers, and can result in more frequent Battery RTL tactics and increased congestion over the launch zone.\footnote{UAV batteries are manually swapped in the launch zone by CCAST personnel in order to support continued mission progress.} 

The CCAST team manages hundreds of batteries, that can vary in age and usage, as well as by vehicle platform type; thus, the CCAST team makes no attempt to develop individual battery specific consumption models for any vehicle type. However, the CCAST multi-resolution simulator needs to integrate battery consumption models. The simulator incorporates a configurable battery life that uses a normal distribution to assign virtual vehicles a battery power level upon deployment. These useful battery durations are specified by virtual vehicle type (e.g., Solos: 22 minutes, M500s: 30 minutes). The virtual vehicles' battery consumption is based on a linear battery model. While the simulator's battery consumption models differ from the hardware vehicle usage, especially for UAVs, it provides a sufficient proxy to support the presented congestion analysis.  

\section{Background} \label{sec:background}
The possibility of using robot swarms to conduct complex missions in built environments has become increasingly relevant \cite{chung_live_fly,chung_survey,skorobogatov_survey}. Recurring swarm related challenges include congestion management, GPS error, detection and avoidance of other vehicles, and difficulties in managing large-scale UAV takeoffs and landings. 

Early decentralized congestion management research focused on leveraging road networks and operating a network of automobiles without intersection signals. Methods ranging from using rule sets  \cite{grossman_traffic_88} to automobiles driving in patterns \cite{ikemoto_intersections} were evaluated. More recent efforts used auction based techniques to manage intersection crossings \cite{carlino_auction_traffic_control}. Each of these methods mitigated intersection congestion and collisions successfully, but was specifically tailored to roadway environments, where the vehicles drive in designated lanes and follow common intersection crossing standards. While the OFFSET FX CACTFs' have road networks that CCAST leverages for UGV navigation, the UGVs are much smaller than automobiles. Two issues arise. First, the GPS localization error for the CCAST UGVs is large, up to 5m, compared to using GPS with automobiles that generally do not encounter significant localization errors. Second, optimizing the mission execution seeks to deploy multiple UGVs on the road network simultaneously, such as multiple UGVs navigating as a group to a goal location. This type of UGV deployment is not required, nor does it need to follow the roadway usage rules applied to automobiles. The OFFSET missions are intended to present a dynamic environment in which the vehicles are not constrained to road networks, with target locations being potentially anywhere in the CACTF, including in fields, open areas between buildings, and even inside buildings. As well, the CCAST UAVs do not have to follow the roadways at all, but do have to navigate while avoiding other UAVs, structures, and obstacles in the CACTF environment. As a result, these road network based methods are unsuitable proxies for analyzing OFFSET relevant congestion scenarios, particularly when focused on UAVs.  

Swarms deployed with unconstrained areas of operation can also encounter congestion \cite{lerman2002_interference}. This effort demonstrated that overall swarm performance increased with the swarm's size, but that individual vehicle performance was shown to degrade as the total number of deployed vehicles increased. However, as size increased further, performance diminished and eventually resulted in negative returns. These results suggest that as the CCAST swarm size increases, the vehicles are expected to increasingly interfere with each other, resulting in congestion reduction becoming an increasingly relevant consideration.

Pareto optimal path planning was explored as a solution to the interference problem \cite{inalhan2002decentralized,ghrist2004pareto}, as were sub-optimal centralized planning methods \cite{turpin_multi_agent_paths,saha_multi_agent_paths}, and dynamic models that predicted and reacted to neighboring UAVs' actions during flight~\cite{senthil_lswarm,arul_dcad}. A limitation of these methods is the relatively few dozen vehicles to which they were applied and their inability  to scale to the required OFFSET swarm size. Further, these methods were designed for a homogeneous swarm performing a single task. The OFFSET mission scenarios require a heterogeneous swarm simultaneously executing a diverse set of tactics. 

Centralized algorithms can provide organized UAV swarm landings~\cite{thomas_centralized_land_paths,dono_centralized_land_paths}, or sequenced aerial holding pattern zones from which UAV's landings are executed using a follow the leader approach~\cite{nazarov_buffers_zone_landing}. While the CCAST vehicles report their telemetry to a centralized process (i.e., the dispatcher), and that telemetry is shared with the swarm's vehicles, each vehicle conducts on-board decentralized navigation path planning~\cite{ccast}. While a centrally coordinated tactic is feasible within the CCAST system, this approach is not preferable when deploying decentralized swarm vehicles. A CCAST objective is for vehicles to conduct their mission assignments until the minimum safe battery level is reached. Once that battery level is reached, a vehicle RTLs. An aerial buffer zone, such as that required for following the leader landings, will require vehicles, particularly UAVs, to have an additional reserve power threshold that is higher than the current system requirements. A higher reserve power threshold will further reduce UAVs' time-on-task and can reduce the swarm's overall performance. Finally, none of these solutions incorporate simultaneous UAVs launching and landing from the same launch zone, which occurs when previously launched UAVs RTL at the same time UAVs takeoff based on newly issued tactics.

Probabilistic state machines were explored to serve as a congestion reduction methodology \cite{marcolino2009traffic}. The state machines require vehicles to randomly wait in close proximity to a target in order to avoid congestion. An extension created pie-shaped ingress and egress lanes to the target \cite{soriano2017avoiding}. Requiring UAVs to wait randomly while hovering at altitude expends more battery than en-route flight and will necessitate allocating additional reserve power with an increased threshold to trigger the CCAST Battery RTL tactic. The use of lanes around a target works for a singular target situation, but likely will not generalize to the OFFSET domain. The OFFSET mission objectives often incorporate multiple targets that can result in overlapping lanes and may cause entire regions to become inaccessible. A potential (i.e., untested) probabilistic state machine variant relevant to CCAST can have the dispatcher assign a random wait time to UAVs just prior to their launch. This approach avoids UAVs hovering in the air unnecessarily and may potentially reduce congestion without lowering the Battery RTL threshold or sacrificing additional battery life.

Coordinated UAV takeoffs can reduce congestion by minimizing the time for swarm UAVs to launch and create pre-designated aerial formations, which require direct UAV-to-UAV communication \cite{fabra_takeoff,hernandez_takeoff}. These formation-forming methods create a localized sub-swarm of co-located UAVs; however, the CCAST mission plan frequently deploys multiple smaller sub-swarms with assigned tactics that are distributed throughout the CACTF. The OFFSET program specifies a single launch zone, but places no requirements on vehicle recovery, meaning vehicles are not required to return to the launch zone. While CCAST can support distributed landing zones, UAV battery replacement necessary to support ongoing mission tempo can only be achieved by human teammates. Therefore, CCAST's operating procedure generally assumes UAVs return to the same location from which they launched within the launch zone. Coupling this procedure with the mission's tempo can result in UAVs launching from and returning to the launch zone at irregular intervals. It is also not uncommon to have UAVs with a completed tactic (e.g., a Building surveil) RTL at the same time new UAVs are launched to address a new tactic. 

Entertainment swarm light shows launch thousands of UAVs. The UAVs are typically placed in rows, and the light show choreography launches them by alternating which rows takeoff as waves~\cite{drone_show_video}. These highly choreographed light shows are generally conducted at AGLs that place the UAVs high above structures and obstacles. The UAVs' actions are typically quite simplistic, often relying on pre-programmed deconflicted navigation paths, especially compared to common CCAST tactics conducted in the dense CACTF air space. The OFFSET mission objectives generally are vastly more complex, requiring large numbers of UAVs to takeoff from significantly smaller launch zones to conduct tactics that require dispersed navigation path plans across the CACTF. The CCAST swarm relies on the tactic's specified goal location's proximity (e.g., the building's location for a surveil) to assign the spatially closest vehicles automatically. The assigned vehicles each individually plan their navigation path to the respective locations at which the tactic is executed. 

The relationship between robot size, robot quantity, and congestion was explored recently \cite{schroeder2019balancing}. Specifically, the balance between the total swarm cost, as a function of robot size and quantity, and interference between vehicles was used to identify the optimal physical size of robots that comprise a swarm. The results rely on the positive correlation between the robots' physical size and their performance. The CCAST's UAVs' performance is less dependent on their physical size, as improved CCAST vehicle performance for the OFFSET domain generally comes at the cost of higher quality sensor payloads.

An immediate swarm congestion concern is UAV battery drainage prior to tactic completion, particularly for persistent tactics. Automatic UAV battery recharging is feasible \cite{erdelj_uav_fly_forever,svogor_battery}, but is beyond the current CCAST swarm system capabilities. The CCAST swarm uses a \textit{swap} algorithm in which UAVs automatically transfer their tactic to another UAV with a full battery \cite{DiehlDARS2022}. Two types of swaps are achieved. UAVs conducting persistent tactics request a replacement UAV and remain on task until the new UAV arrives, as allocated by the dispatcher. UAVs performing interruptible tasks, relinquish their tactic to the dispatcher and execute the RTL behavior. The dispatcher selects a replacement UAV that launches to continue performing the prior UAVs' tactic. These approaches can address the ``symptoms'' of congestion, but can also create additional traffic that may increase congestion.

Swarm congestion can occur for many reasons. The complexity of the CCAST swarm, in conjunction with the DARPA OFFSET mission deployment constraints introduce new factors that will impact swarm congestion. Conducting the mission effectively and preparing a feasible multiple phase mission plan requires understanding how all the facets of the mission, the CCAST system, and constraints, such as launch zone space limitations impact congestion during a mission deployment. None of the existing literature addresses all the constraints encountered during the DARPA OFFSET program. 

\section{Pre-FX6 Launch Zone Configuration Analysis}

The maximum number of vehicles that can be deployed at a CACTF is determined by multiple constraints, such as the DARPA designated launch zone area, environmental obstacles (e.g., trees and power lines), and the size of each vehicle's GPS error-associated safety zones. The CCAST architecture implements a \textit{safety distance} in order to avoid vehicles unnecessarily colliding with one another when departing from the launch zone. The safety distance for all UAVs is 1m, while the UGV distance is 3m. The difficulty is that given the CCAST vehicle's sizes, their dimensions are provided in Table~\ref{Tab:Robots}, and their GPS systems, the GPS error can be up to 5m. The minimum safe operation distance between two vehicles is the sum of the vehicles' safety distances (e.g., 2m between two UAVs, 4m between a UAV and a UGV).

The CCAST team must use the DARPA specified launch zone; therefore, a key initial question is how many UGVs and UAVs can fit within the launch zone, while also meeting the CCAST specified safety distances? The planning for FX6 assumed that 240 vehicles (40 UGVs, 40 3DR Solos, 20 Uvify IFO-Ss, and 140 VOXL m500s UAVs) had to be safely accommodated within the FX6 launch zone. Early in the FX planning, the Leschi Town CACTF at Joint Base Lewis McChord was the intended FX6 destination; however, later the location changed to Fort Campbell's Cassidy CACTF. Analyses for both CACTFs are reported.

\subsection{Joint Base Lewis McChord, Leschi Town CACTF} \label{sec:jblm_theoretical_analysis}
\begin{figure*}[!htb]
    \centering
    \includegraphics[width=\textwidth]{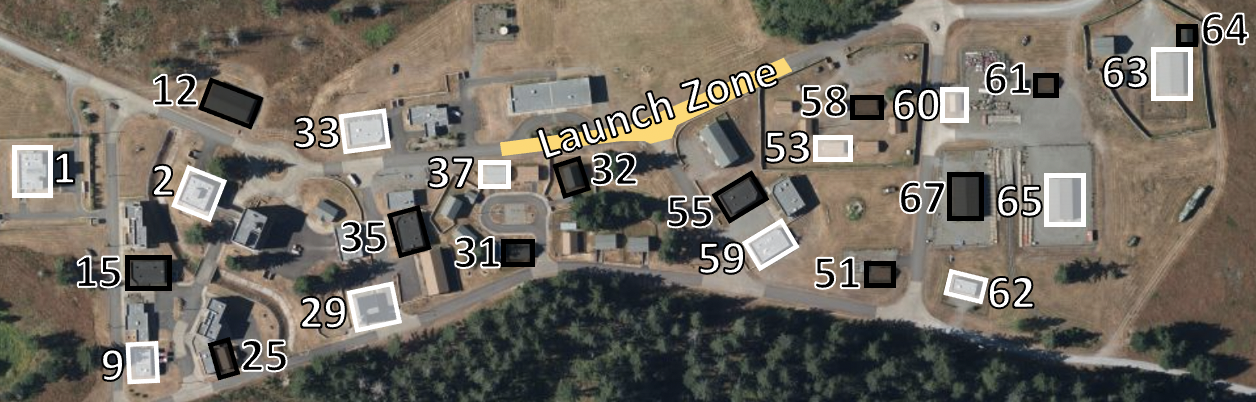}
    \caption{The Joint Base Lewis McChord Leschi Town CACTF showing the buildings by Building set (A: white and B: black) and the launch zone (yellow). }
    \label{fig:jblm_map_numbered}
\end{figure*}

Joint Base Lewis McChord's Leschi Town CACTF, see Figure \ref{fig:jblm_map_numbered}, is roughly 200,000 meters\textsuperscript{2} (m$^{2}$), approximately 250m north-to-south x 800m east-to-west. Fifty one to five story buildings are dispersed throughout the CACTF, which also includes light posts, street signs, natural vegetation and trees, drainage ditches, barricades, a playground, etc. The planned primary launch zone, shown in Figure \ref{fig:jblm_map_numbered}, was a 7.5m wide x 170m long section of roadway (1300m$^2$) with grass or curb borders. 

A 3 x 80 vehicle configuration with a 2m spacing between UGVs and a 1.5m spacing between UAVs permits 240 vehicles to fit within the launch zone; however, this configuration does not account for any GPS error, or the minimum safety distances required for safe swarm operation. Given CCAST's defined safety distances, the minimum safe operating distance between two vehicles is determined to be 2m between two UAVs, 6m between two UGVs, and 4m between a UAV and a UGV. 
Adherence to the minimum safety distance between vehicles, while maximizing the number of vehicles that can fit into the launch zone size results in a maximum of 18 UGVs and 112 UAVs arranged in two rows of 65 vehicles each. However, this configuration falls 110 vehicles short of the intended goal swarm size.  

\subsection{Fort Campbell, Cassidy CACTF}  \label{sec:cassidy_theoretical_analysis}
\begin{figure*}[!h]
    \centering
    \includegraphics[width=0.49\textwidth]{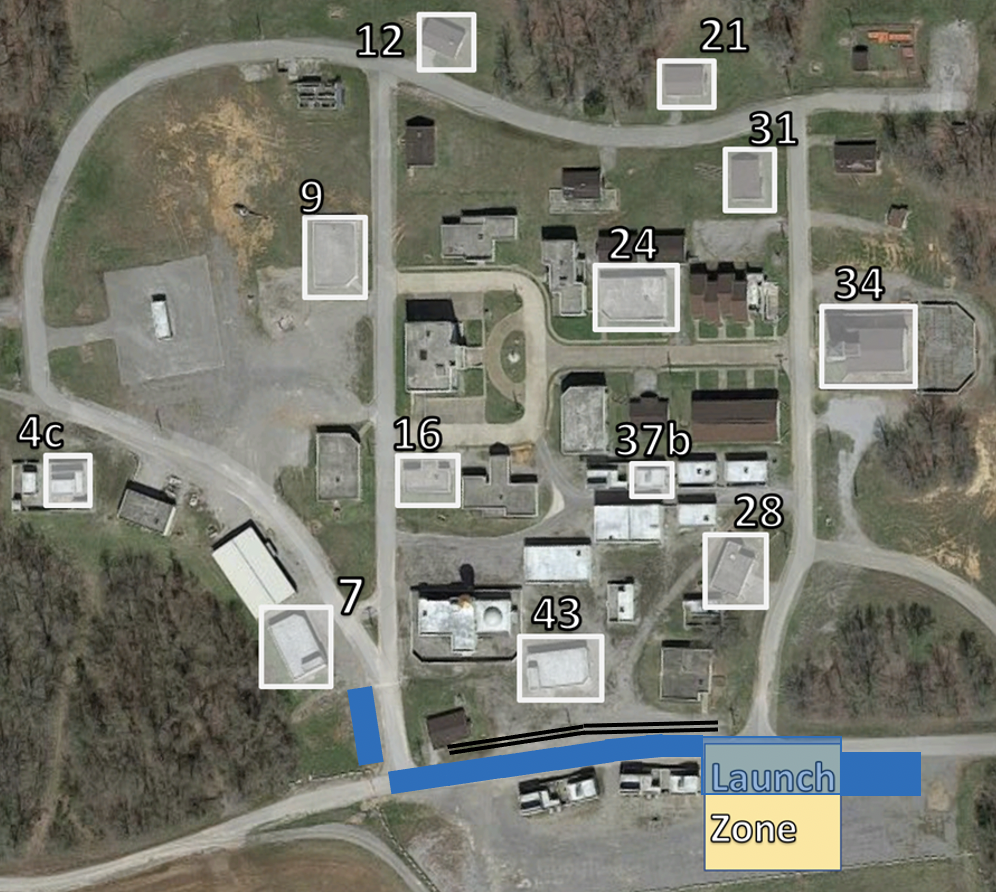}
    \caption{The Fort Campbell Cassidy CACTF showing the tested building set, the pre-FX expected and analyzed launch zone (yellow), the actual FX6 launch zone (blue areas), and the approximate location of power lines close to the launch zone (black lines). }
    \label{fig:cassidy_map_numbered}
\end{figure*}

Ultimately, FX6 was held at the Fort Campbell Cassidy CACTF, see Figure \ref{fig:cassidy_map_numbered}. This CACTF is roughly 100,000 m\textsuperscript{2}, approximately 350m north-south x 285m east-west, with 43 one to five story buildings that are more densely distributed than the Leschi Town CACTF. The pre-FX launch zone, shown as yellow in Figure \ref{fig:cassidy_map_numbered}, used for the presented pre-FX6 analysis, is an approximately 37m north-south x 41m east-west area primarily covering a parking lot and a small portion of the roadway (1517 m\textsuperscript{2}). The actual FX launch zone, shown as the blue areas in the figure, was roughly the same total size (1468m\textsuperscript{2}), but had a different spatial distribution. The small launch area on the left, close to the building labeled 7, is approximately 6m wide x 16m long. The roadway between the buildings was the largest continuous area, measuring approximately 6m wide x 142m long. The smaller area to the right of the buildings, which overlaps with the anticipated Pre-FX launch zone, is 10m wide x 52m long. All launch areas' ground-based borders were generally grass, gravel, or pavement; however, power lines, shown in their approximate position as the black lines in the figure, hugged the roadway and impacted UAV launch zone placements. 

A 15 row x 16 column vehicle configuration with a 2.5m spacing allows 240 vehicles to fit in the designated launch zone. However, as with the Leschi Town analysis, this configuration does not account for GPS error or the minimum safety distance requirements, which were violated for the UGVs. Adherence to the safety distance requirements between the vehicles results in a maximum of 14 rows. 12 rows with 16 columns are allocated to the UAVs, and the UGVs had a 2 row x 7 column placement. This configuration accommodates 192 UAVs and 14 UGVs, which is 34 vehicles short of the 240 vehicle goal.

An analysis of the launch zones for both the Leschi Town and Cassidy CACTFs  provided insufficient space to deploy the entire planned swarm and maintain the CCAST safety distances. Further, prior field exercises demonstrated that a launch zone spacing twice the minimum specified safety distances was often necessary to avoid congestion. Thus, richer analyses of inter-vehicle spacing, the potential of using deployment waves, the vehicle placement pattern within the launch zone, and the potential impact of the mission plan composition were identified as additional alternatives for safely deploying 240 vehicles from the limited launch zone.

\section{Pre-FX6 Congestion Analysis} \label{sec:pre_fx}

The largest impact from congestion, given the CCAST swarm's composition, is on the UAVs; thus, the Pre-FX6 analysis focus on them, and the UGVs are excluded. The analysis hypotheses focus on understanding how the impacts of vehicle placement spacing, using deployment waves, and vehicle placement patterns can decrease congestion. 
\textit{Hypothesis \RomanNumeralCaps{1}} states that increasing launch zone spacing will decrease swarm congestion.
\textit{Hypothesis \RomanNumeralCaps{2}} states that increasing the number of waves will decrease swarm congestion.
\textit{Hypothesis \RomanNumeralCaps{3}} states that using a hexagonal layout in the UAV launch zone, as opposed to a square layout, will use less space without increasing congestion. 
While it may appear to the causal reader that the first two hypotheses are obvious, given the FX6 and CCAST system constraints, that is not necessarily true. \textit{Hypothesis \RomanNumeralCaps{1}} and \textit{\RomanNumeralCaps{2}} are evaluated using the experiments presented in  Sections~\ref{sec:pre_fx_jblm} and \ref{sec:pre_fx_cassidy}. The third hypothesis is loosely based on prior FX launch zone trial and error patterns, which is a focus of the Section~\ref{sec:pre_fx_cassidy} experiment. 

\subsection{Pre-FX6: Joint Base Lewis McChord, Leschi Town CACTF} \label{sec:pre_fx_jblm}
The Pre-FX6 Leschi Town CACTF evaluation incorporated mission plans composed of multiple Building surveil tactics, but varied the UAV launch zone spacing (\textit{Hypothesis \RomanNumeralCaps{1}}) and the number of waves in which the UAVs are deployed (\textit{Hypothesis \RomanNumeralCaps{2}}). This experiment was conducted using the CCAST swarm architecture and associated multi-resolution simulator. Two mission plans were developed that each incorporated twelve Building surveil tactics. Each Building surveil tactic required five UAVs, four with a forward-facing camera payload and the fifth with a downward-facing camera payload. A  total of 60 UAVs were required to complete these mission plans, 48 with forward-facing and 12 with downward-facing camera payloads. 

\subsubsection{Experimental Methodology}

\paragraph{Independent Variables} \label{sec:jblm_indp_vars}
All experimental trials required a mission plan that incorporated relevant tactics. The developed mission plans incorporate two sets of twelve buildings distributed across the CACTF, as identified in Figure \ref{fig:jblm_map_numbered}. The selected buildings and resulting mission plans represent independent variables. The general mission plan focused on the Phase I intelligence gathering aspects of a typical FX mission, and; therefore, focused on issuing Building surveil tactics to collect available intelligence information.  The building sets were used to generate multiple mission plans, whose creation will be explained.

The remaining independent variables focus on the swarm vehicles' placement and usage of the available launch zone area. The \textit{launch zone spacing}, or the distance between UAVs, was varied between 2m and 5m in 1m increments. The total \textit{number of waves} focused on how many launch waves a mission plan contained. A single (1) wave launched all UAVs simultaneously. The remaining evaluated number of waves values divided the UAVs into groups based on the number of waves the mission plan contained, either 2, 3, 4, or 6. 

The \textit{number of regions} was dependent on both the number of buildings and the number of waves and decreases as the number of waves increases. The number of regions evaluated (i.e., 12, 6, 4, 3 and 2) was calculated as follows: $\textit{number of regions} = \textit{number of buildings} / \textit{number of waves}$. A mission plan incorporating 12 buildings and 6 waves results in 2 regions. Adding buildings, not varied in these evaluations, or fewer waves result in a larger number of regions. A qualitative analysis determined that 90 seconds (s) between waves was sufficient, as it generally allows the current wave of UAVs to launch and begin moving towards their goal before the next wave was tasked. If the time between waves decreases, then the likelihood of congestion increases. While a larger time between waves may reduce the likelihood of congestion, it may also increase congestion if UAVs deployed in earlier waves return to the launch zone at the same time a new wave launches.

The \textit{number of tactic invocations per wave} was similarly dependent on the number of buildings and the number of launch waves. The number of tactic invocations per wave was always equivalent to the number of regions, or a single tactic invocation per region, per wave. Thus, the number of tactic invocations per wave was evaluated using 12, 6, 4, 3, and 2 tactic invocations per wave. While it is possible to have multiple tactics assigned to a region during a wave, such assignments are considered outside the scope of this analysis. 

\paragraph{Dependent Variables} \label{sec:jblm_dep_vars}
The CCAST swarm UAVs launch, ascend to altitude, and must complete their navigation path planning prior to commencing travel to achieve the assigned tactics. Thus, an extensive period hovering at altitude due to congestion and blockage can cause a UAV to consume its battery, resulting in it RTLing without contributing to the mission objectives, a highly undesirable outcome. During the FX, the swarm commander may explicitly attempt to ``assist'' the vehicle in becoming unblocked. Specifically, the swarm commander may Nudge a vehicle, which in the case of a UAV causes it to change altitude slightly, in hopes of unblocking navigable paths. A more severe swarm commander action Stops the UAV's current tactic, which is followed by issuing either an RTL or an entirely new tactic. The swarm commander's options to assist the vehicle in becoming unblocked are not incorporated into the evaluation trials.

An \textit{independent block} occurs when no clear navigation path is available (i.e., a navigation path plan cannot be generated), and the UAV continues to search for a viable path plan. CCAST swarm UAVs mark themselves as blocked immediately when path planning fails, or a mobile object blocks its path. The path planning process resets after a block has persisted for 10 seconds, meaning the vehicle marks itself as unblocked and restarts the path planning process. After the third reset due to blocks (i.e., 30 seconds), the planner resets and the entire path planning process resets. These steps are repeated until either a navigable path is identified, or the UAV's battery reaches the Battery RTL threshold, at which point it will RTL. Consecutive blocks that occur within ten seconds of one another are counted as a single block. Multiple independent blockages can occur for the same vehicle within the same tactic execution, or within a shift, since the vehicle can encounter a new blockage situation as it executes a navigation plan in the environment. 

The amount of time a vehicle is blocked is called the block duration. An \textit{independent block duration} is the amount of time over which an independent block occurs. Multiple consecutive blockages are combined into a single independent block. The corresponding block durations for each of the combined consecutive blockages are summed to create the block duration. The latitude and longitude of each block event are also recorded.

Upon trial completion the \textit{total block count} and \textit{total block duration} is calculated by summing all recorded independent blocks and the independent block durations, respectively. The independent and total blockage durations are measured, in milliseconds, but the total block duration results are reported as minutes.

\paragraph{Mission Plan Design} \label{sec:jblm_plan_design}

The CCAST FX mission plans are developed prior to shift deployments based on information pertaining to the number of available vehicles, their types and payloads, mission objectives and the tactics required to achieve that mission, the launch zone and other environmental constraints, prior intelligence information,  etc. Conducting the evaluations requires developing representative mission plans. Independent mission plans were developed to account for all combinations of the independent variables. Each mission plan focused on the Phase I information gathering mission objective. Specifically, each mission assigned the vehicles to Building surveil tactics. The CCAST team's airspace deconfliction heuristics were applied to selecting the buildings to be surveilled. The selected buildings are depicted in Figure \ref{fig:jblm_map_numbered}.

\textit{Region creation} allocates the buildings to separate CACTF areas (i.e., regions) in order to organize the wave deployments. Region creation is required for each number of waves. The resulting regions are used to generate the corresponding mission plan. At a minimum, each region must contain at least one building to be surveilled. The regions ideally radiate outwards from the launch zone, resembling pie slices if the launch zone was centered in a circular CACTF.

The CCAST architecture can signal multiple tactics (e.g., multiple Building surveils) to be executed simultaneously, which permits launch waves. Once the regions are identified, where each region contains an equivalent number of buildings, waves are assigned to the buildings inside  each region. The building assignment is repeated for each number of waves value. The first wave begins with the outer perimeter of buildings, invoking the furthest Building surveil tactic from the launch zone in each region. Subsequent waves assign buildings to the tactics by moving inwards (e.g., closer to the launch zone) from the last tactic's building assignment, which allows earlier UAV waves to clear the launch zone before the next wave launches. 

Two sets of buildings were used to generate the mission plans, Building set A and B, the buildings are labeled in Figure \ref{fig:jblm_map_numbered}. The specific Leschi Town Building set assignments for a given number of launch waves are provided in Table~\ref{tab:wave_assignments}. Each table entry decomposes the buildings, represented by their number from the figure, into the required number of launch waves. 
Ten mission plans, five per building allocation (i.e., Building Set A and B), were developed. While the swarm commander typically launches the mission plan waves, this experiment used a program script to instantiate the launch waves.

\begin{table}[!h]
\begin{center}
\begin{tabular}{|p{20mm}||p{40mm}|p{40mm}||p{40mm}|}
\hline
\multirow{2}{*}{\textbf{\# Waves}} & \multicolumn{2}{ c||}{\textbf{Leschi Town }} & \textbf{Cassidy} \\ \cline{2-4} 
& \textbf{Building Set A} & \textbf{Building Set B} & \textbf{Building Set}\\ \hline
\textbf{1} & 
\begin{tabular}[c]{@{}l@{}} 1,2,9,29,33,37, \\ 53,59,60, 62,63,65\end{tabular} &
\begin{tabular}[c]{@{}l@{}}12,15,25,31,32,35,\\ 51,55,58,61,64,67\end{tabular} &
\begin{tabular}[c]{@{}l@{}}4c,7,9,12,16,21,\\ 24,28,31,34,37b,43\end{tabular} \\ \hline 
\textbf{2} & 
\begin{tabular}[c]{@{}l@{}}\textbf{1:} 1,9,33,62,63,65; \\ \textbf{2:} 2,29,37,53,59,60\end{tabular} &
\begin{tabular}[c]{@{}l@{}}\textbf{1:} 12,15,25,55,64,67;\\ \textbf{2:} 31,32,35,51,58,61\end{tabular} &
\begin{tabular}[c]{@{}l@{}}\textbf{1:} 4c,12,21,31,34,16;\\ \textbf{2:} 7,9,24,28,37b,43\end{tabular} \\ \hline 

\textbf{3}  & 
\begin{tabular}[c]{@{}l@{}}\textbf{1:} 1,9,63,65; \\ \textbf{2:} 2,29,60,62; \\ \textbf{3:} 33,37,53,59\end{tabular}  &
\begin{tabular}[c]{@{}l@{}}\textbf{1:} 15,25,64,67;\\ \textbf{2:} 12,31,51,61; \\ \textbf{3:} 32,35,55,58\end{tabular}  &
\begin{tabular}[c]{@{}l@{}}\textbf{1:} 4c,12,21,34;\\ \textbf{2:} 7,9,24,31; \\ \textbf{3:} 16,28,37b,43\end{tabular} \\ \hline

\textbf{4} & 
\begin{tabular}[c]{@{}l@{}}\textbf{1:} 1,9,63; \textbf{2:} 2,62,65;\\ \textbf{3:} 29,33,60; \textbf{4:} 37,53,59\end{tabular} & 
\begin{tabular}[c]{@{}l@{}}\textbf{1:} 15,64,67;  \textbf{2:} 12,25,61; \\ \textbf{3:} 35,51,58; \textbf{4:} 31,32,55\end{tabular} &
\begin{tabular}[c]{@{}l@{}}\textbf{1:} 12,21,34;  \textbf{2:} 9,24,31; \\ \textbf{3:} 4c,16,37b; \textbf{4:} 7,28,43\end{tabular}  \\ \hline

\textbf{6}  & 
\begin{tabular}[c]{@{}l@{}}\textbf{1:} 1,63; \textbf{2:} 9,65;\\ \textbf{3:} 2,62; \textbf{4:} 29,59; \\ \textbf{5:} 33,60; \textbf{6:} 37,53\end{tabular} & 
\begin{tabular}[c]{@{}l@{}}\textbf{1:} 15,64; \textbf{2:} 25,67;\\ \textbf{3:} 12,61; \textbf{4:} 35,51;\\ \textbf{5:} 31,58; \textbf{6:} 32,55\end{tabular} &

\begin{tabular}[c]{@{}l@{}}\textbf{1:} 12,21; \textbf{2:} 4c,34;\\ \textbf{3:} 9,31; \textbf{4:} 16,24;\\ \textbf{5:} 7,37b; \textbf{6:} 28,43\end{tabular}  \\ \hline

\end{tabular}
\end{center}
\caption{The Leschi Town and Cassidy CACTFs' (A and B) Building set assignments by number of launch waves (\# Waves).}
\label{tab:wave_assignments}
\end{table}

\paragraph{Launch Zone Configuration}
The CCAST multi-resolution swarm simulator requires a launch zone configuration file for each mission plan. This configuration file defines the vehicles, their types, and their launch/home locations within the launch zone. Separate launch zone configuration files must be created for each launch zone spacing value. The launch zone was configured into two rows of 30 UAVs along the road, based on the results from Section \ref{sec:jblm_theoretical_analysis}.

\paragraph{Experiment Execution}
The computational complexity of the experimental design will vary depending on the specific swarm simulator, the number of vehicles in the swarm, and the mission plan complexity. The analysis of the experimental results are dependent on the complexity of the produced log files and the amount of generated data to be analyzed.

Twenty trials was performed for each combination of launch zone spacing and number of waves. Each number of waves has an independent mission plan, resulting in five mission plans per Leschi Town Building set (A and B). 800 trials were executed, 400 per Building set (i.e., 5 mission plans x 4 launch zone configurations x 20 trials x 2 building sets = 800). After the final wave launched, each simulation trial ran for 20 minutes. Prior analysis of the UAV Swap tactic demonstrated that the average 3DR Solo battery life, the lowest of all CCAST UAVs, was under 20 minutes \cite{DiehlDARS2022}.

\subsubsection{Results} \label{sec:jblm_results}

\begin{figure*}[!htb]
\centering
\begin{subfigure}{.90\textwidth}
    \centering
    \includegraphics[width=\linewidth]{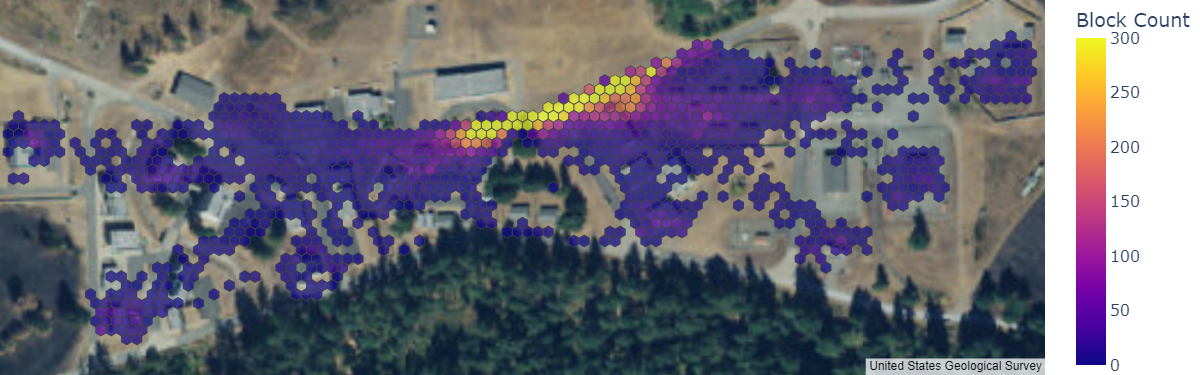}
    \caption{Heatmap of the location of congestion instances. }
    \label{fig:jblm_overall_map}
\end{subfigure}
\hfill
\begin{subfigure}{.55\textwidth}
    \centering
    \includegraphics[width=.9\linewidth]{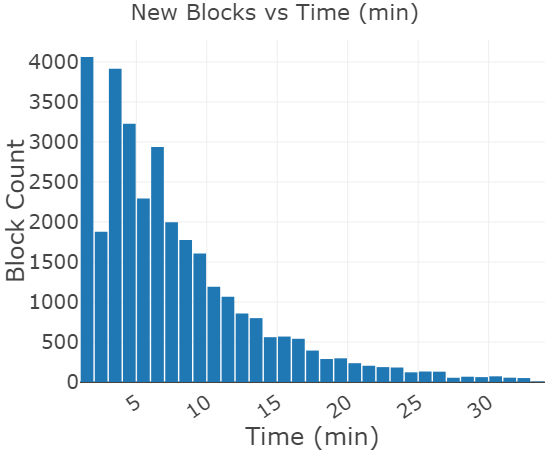}
    \caption{Histogram of the block start times.}
    \label{fig:jblm_a_all_blocks_histogram}
\end{subfigure}
\caption{Joint Base Lewis McChord, Leschi Town CACTF congestion instance location heatmap across all Building set A simulations (a), and times block instances began histogram by single minute increments (b).}
\end{figure*}

An overall heatmap of all locations at which congestion occurred across all simulations for Building set A, as shown in Figure~\ref{fig:jblm_overall_map}, can be used to identify the locations of potential congestion. Across the Building set A trials, congestion occurs across the CACTF, as shown in the figure. While there is some increased congestion near the buildings specified for this set, the vast majority of the congestion occurs along the launch zone (yellow and orange in the figure). The heatmap generated for Building set B had a similar distribution of block locations, with the dominant locations being the launch zone, followed by the buildings in the set. 

A histogram of each blocks' start times is provided in Figure~\ref{fig:jblm_a_all_blocks_histogram}. 
The majority of blockages occurred during the initial UAV wave deployments. Recall that simulation trials containing multiple waves launch additional waves at 90 second intervals (i.e., $1^{st}$ wave: 0 minutes, $2^{nd}$: 1.5 minutes, $3^{rd}$: 2 minutes, $4^{th}$: 4.5 minutes, and $6^{th}$: 7.5 minutes). The histogram demonstrates that the majority of the blocks began when the mission plan's UAV waves launch, after which the number of new blocks is much lower. Once the vehicles launched, the remaining block instances are related to longer duration blockages in the launch zone, blockages out on the CACTF near the buildings to be surveilled, or when the vehicles return to the launch zone.  

\begin{figure*}[!hbt]
\centering
\begin{subfigure}{.5\textwidth}
    \centering
    \includegraphics[width=\linewidth]{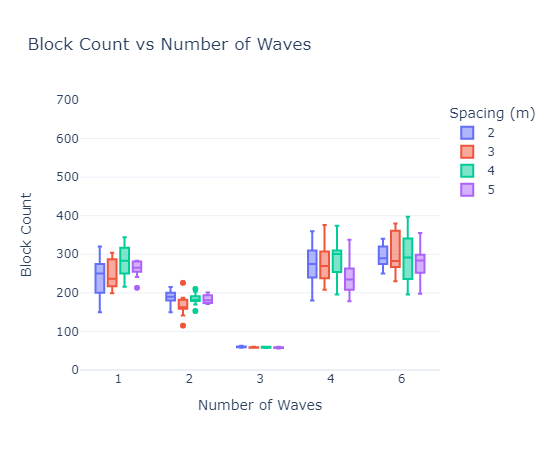}
    \caption{Building set A - block count.}
    \label{fig:jblm_total_count_a}
\end{subfigure}%
\hfill
\begin{subfigure}{.5\textwidth}
    \centering
    \includegraphics[width=\linewidth]{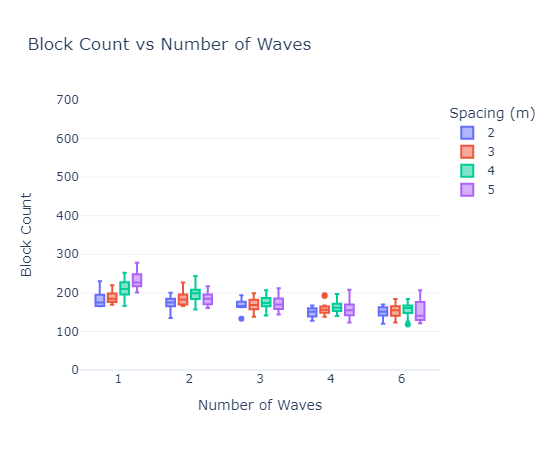}
    \caption{Building set B - block count.}
    \label{fig:jblm_total_count_b}
\end{subfigure}
\begin{subfigure}{.5\textwidth}
    \centering
    \includegraphics[width=\linewidth]{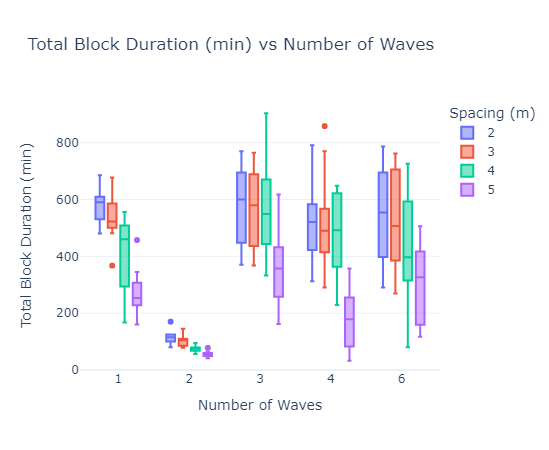}
    \caption{Building set A - block duration.}
    \label{fig:jblm_total_dur_a}
\end{subfigure}%
\hfill
\begin{subfigure}{.5\textwidth}
    \centering
    \includegraphics[width=\linewidth]{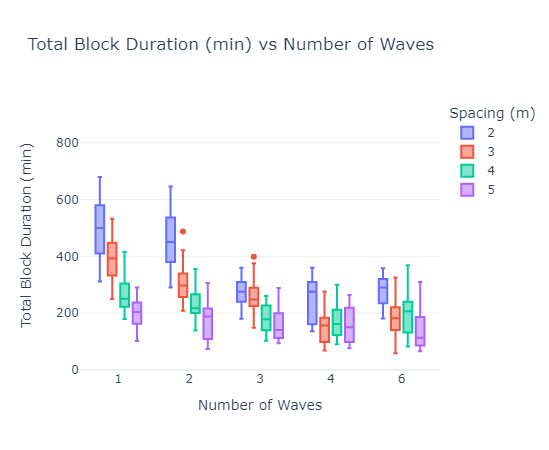}
    \caption{Building set B - block duration.}
    \label{fig:jblm_total_dur_b}
\end{subfigure}
\caption{Joint Base Lewis McChord, Leschi Town CACTF's congestion median total block count and duration box plots by building set, number of launch waves, and launch zone spacing.}
\label{LeschiBoxGraphs}
\end{figure*}

\begin{figure}[!h]
\centering
\begin{subfigure}{.90\textwidth}
    \centering
    \includegraphics[width=\linewidth]{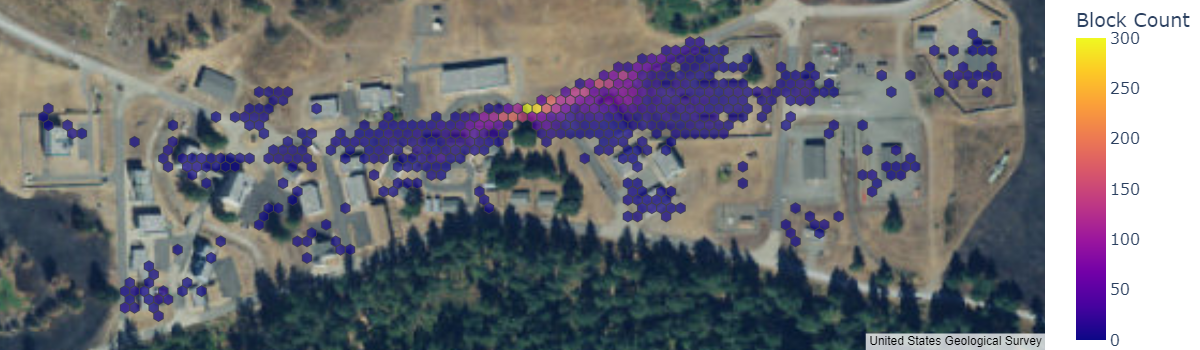}
    \caption{One deployment wave.}
    \label{fig:jblm_a_1_wave}
\end{subfigure}
\hfill
\begin{subfigure}{.90\textwidth}
    \centering
    \includegraphics[width=\linewidth]{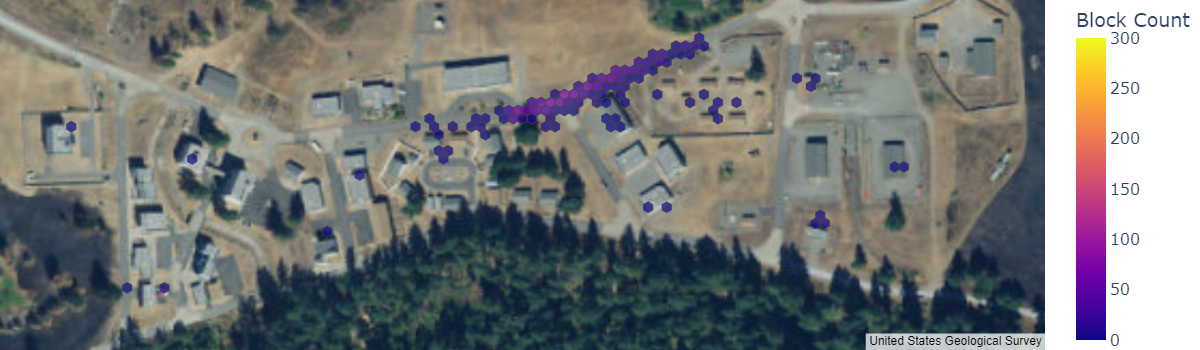}
    \caption{Three deployment waves.}
    \label{fig:jblm_a_3_wave}
\end{subfigure}
\hfill
\begin{subfigure}{.90\textwidth}
    \centering
    \includegraphics[width=\linewidth]{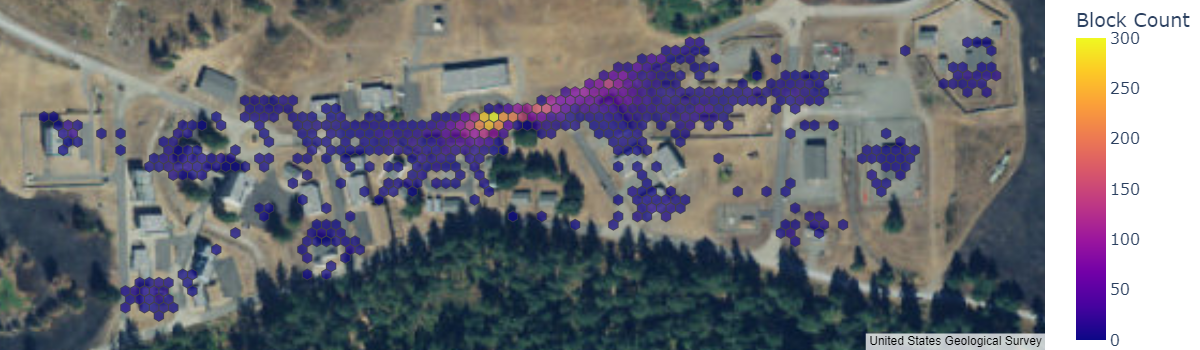}
    \caption{Six deployment waves.}
    \label{fig:jblm_a_6_wave}
\end{subfigure}

\caption{Joint Base Lewis McChord, Leschi Town CACTF's Building set A's congestion heatmap for 1 (a), 3 (b), and 6 (c) deployment waves.}
 \label{fig:jblm_wave_differences}
\end{figure}

The median block counts and block durations for both building sets by number of waves and launch zone spacing are provided in Figure~\ref{LeschiBoxGraphs}. The Building set A 
results shown in Figure~\ref{fig:jblm_total_count_a} indicate that 3 waves, regardless of launch zone spacing, result in the fewest blocks, with 2 waves having the second lowest block count.  Regardless of the launch zone spacing, the number of blocks increased with 4 and 6 waves. 
Overall, Building set A's total block count significantly decreased with 2 and 3 deployment waves, but increased with $>$4 waves. Heatmaps for Building set A's 1, 3, and 6 wave results across all spacings are provided in Figure 7. The 3 wave block count (Figure~\ref{fig:jblm_a_3_wave}) is substantially lower than that of the 1 and 6 wave results (Figure~\ref{fig:jblm_a_1_wave} and c, respectively).  
A 5 (number of waves) $\times$ 4 (launch zone spacing) between-groups ANOVA was performed for Building set A's total block count results.  No significant main effect for the launch zone spacing was found, but a significant main effect existed for the number of waves ($F(4,12) = 189.82, p < 0.01$).  A posthoc Tukey test (p = 0.05) of the pairwise differences by the number of waves found that the 2 and 3 wave instances were both significantly lower than the 1, 4, and 6 wave instances. Additionally, the block count for the 3 wave instances was significantly lower than the 2 wave results.

Generally, little difference existed in Building set A's median total block durations with the 2m, 3m, and 4m spacings across 1, 3, 4 and 6 waves, as depicted in Figure~\ref{fig:jblm_total_dur_a}. However, increasing the launch zone spacing to 5m lowered the total block duration for any number of waves compared to the other spacings. Two waves  had shorter block durations irrespective of spacing, with the tightest minimum and maximum ranges. A 5 (number of waves) $\times$ 4 (launch zone spacing) between-groups ANOVA yielded significant main effects for number of waves ($F(4,12)=44.25, p<0.01$), and launch zone spacing ($F(3,12)=33.97, p<0.01$). A posthoc Tukey test of the pairwise differences by number of waves found that the 2 wave instances were significantly lower than the 1, 3, 4, and 6 wave instances. A posthoc Tukey test assessing differences by launch zone spacing indicated that 5m spacing had a significantly lower block duration than the 2, 3, and 4m spacings. The block duration for 4m was also significantly lower than the 2m spacing.

\begin{figure}[!htbp]
\centering
\begin{subfigure}{.85\textwidth}
    \centering
    \includegraphics[width=\linewidth]{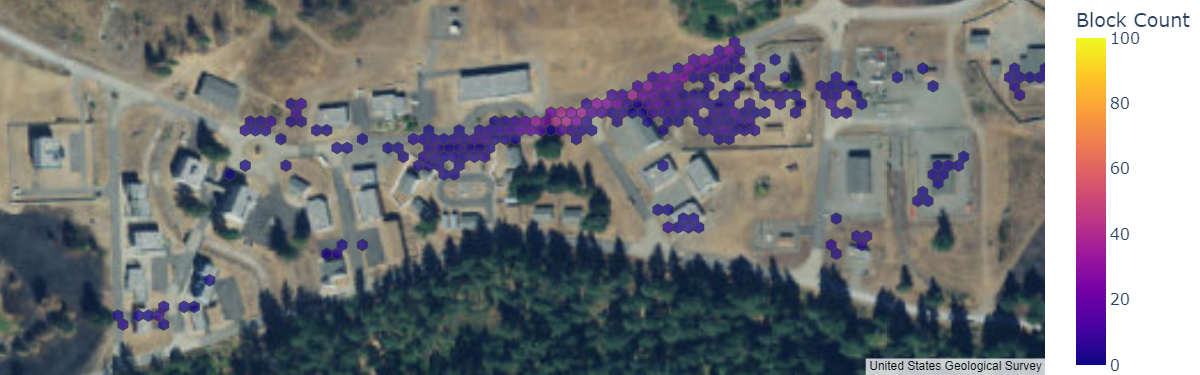}
    \caption{Best-case configuration - block count.}
    \label{fig:jblm_best_count_map}
\end{subfigure}
\hfill
\begin{subfigure}{.85\textwidth}
    \centering
    \includegraphics[width=\linewidth]{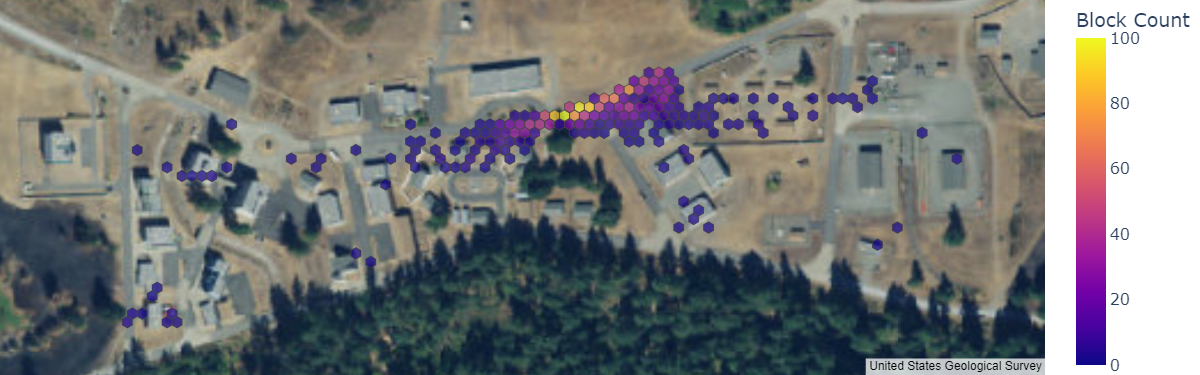}
    \caption{Worst-case configuration - block count.}
    \label{fig:jblm_worst_count_map}
\end{subfigure}%

\begin{subfigure}{.85\textwidth}
    \centering
    \includegraphics[width=\linewidth]{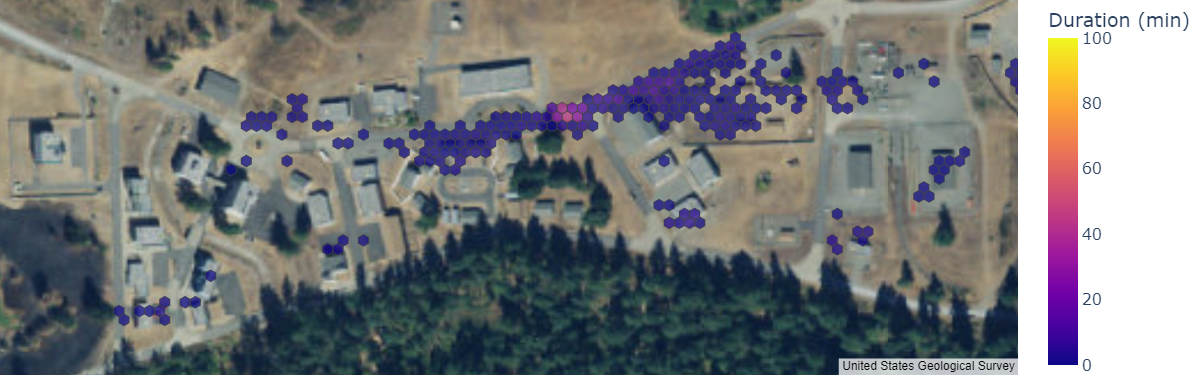}
    \caption{Best-case configuration - block duration.}
    \label{fig:jblm_best_duration_map}
\end{subfigure}
\hfill
\begin{subfigure}{.85\textwidth}
    \centering
    \includegraphics[width=\linewidth]{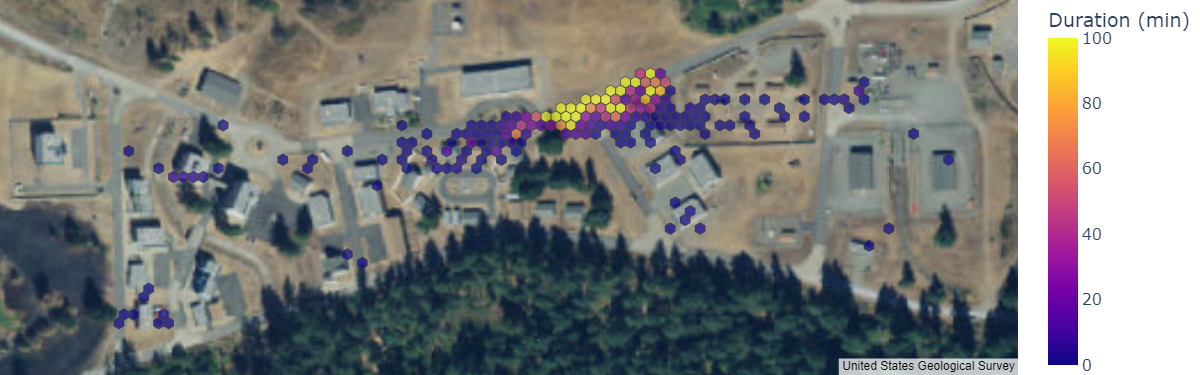}
    \caption{Worst-case configuration - block duration.}
    \label{fig:jblm_worst_duration_map}
\end{subfigure}%
\caption{Joint Base Lewis McChord, Leschi Town CACTF's Building Set A's congestion heat map by best-case (5m Spacing and 2 Waves) total block count (a) and total block duration (b) and worst-case (2m Spacing and 1 Wave) total block count (c) and total block duration (d).}
\label{fig:jblm_best_worst_maps}
\end{figure}

The best- and worst-case configurations for Building Set A  were identified. The best-case occurred for the 5m spacing  with 2 launch waves, while the worst-case had a 2m spacing for 1 Wave. Heatmaps for these configurations’ total block count and total block durations are shown in Figure 8. These heatmaps highlight similarities between the concentrated locations of the total block count and total block duration metrics, where the concentration of more blockages and the longest blockages tend to occur in the launch zone. The worst-case, 2m spacing with 1 launch wave, clearly has more (Figure~\ref{fig:jblm_worst_count_map}) and longer (Figure~\ref{fig:jblm_worst_duration_map}) blockages. Noticeably, the block duration metric more clearly shows the severity of the congestion differences between the best- and worst-cases. The worst-case’s increased blockage counts and durations appear to reduce congestion throughout the CACTF, but this is likely due to the launch zone congestion blocking vehicles from moving out to navigate around the CACTF as required by the assigned tactics. 

\begin{figure}[!htbp]
\centering

\begin{subfigure}{.49\textwidth}
    \centering
    \includegraphics[width=\linewidth]{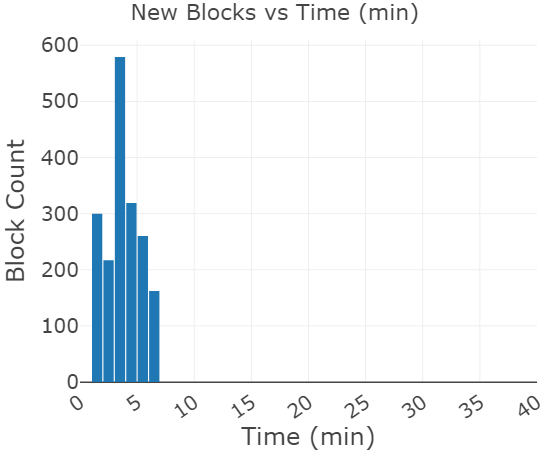}
    \caption{Best-Case configuration - block start time.}
    \label{fig:jblm_best_histogram}
\end{subfigure}%
\hfill
\begin{subfigure}{.49\textwidth}
    \centering
    \includegraphics[width=\linewidth]{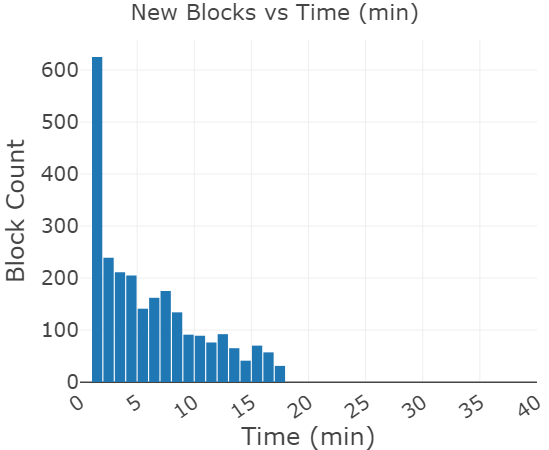}
    \caption{Worst-Case configuration - block start time.}
    \label{fig:jblm_worst_histogram}
\end{subfigure}%

\caption{Joint Base Lewis McChord, Leschi Town CACTF histograms showing the block event start times for the best- (a), and worst-case (b).}
\label{fig:jblm_best_worst_histograms}
\end{figure}

The start times, represented as minutes from the beginning of the mission, of these best-case and worst-case configuration blockages are shown with histograms in Figure~\ref{fig:jblm_best_worst_histograms}. The best case configuration with two launch waves demonstrates that all blockages occur early in the mission. While there are blockages after the first wave launches (i.e., 1-2 minutes), the number of blockages increases substantially after the second wave launches (i.e., 2-3 minutes). The number of new blockages drops, until no new blockages are detected after the $6^{th}$ minute. The worst-case configuration's single launch wave immediately results in the largest number of new blocks (i.e., 1-2 minutes). While the number of new blocks decreases substantially after the first two minutes, recall that the worst-case block durations are much longer than the best-case configuration (as shown in Figure~\ref{fig:jblm_best_worst_maps}). These longer block durations clearly result in new blockages for an extended period into the mission. Whereas the best-case's new blocks across the CACTF occur early in the mission, the worst-case's instances occur throughout the first 18 minutes. 

The Building set B results demonstrate little effect from varying the launch zone spacing, as shown in Figure~\ref{fig:jblm_total_count_b}. The 1 and 2 waves results with 2 and 3m spacings were generally lower than the 4 and 5m spacings. Increasing the number of waves consistently decreased the total block counts. Heatmaps by the number of waves, similar to Figure~\ref{fig:jblm_wave_differences} were not included to conserve space. However, the heatmaps also reflect the decline with increased wave size, but are otherwise similar to Building set A, with the majority of blocks occurring in the launch zone, with some higher incidences of blocks near the set's buildings.
The 5 (number of waves) $\times$ 4 (launch zone spacing) between-groups ANOVA for the total block count values yielded significant main effects for the number of waves ($F(count) = 88.44, p < 0.01$), and launch zone spacing ($F(4,12) = 7.92, p < 0.01$). A posthoc Tukey test (p = 0.05) assessing differences by the number of waves
found that there was no significant difference between the 4 and 6 wave instances, but all other instances were significantly different. A posthoc Tukey test (p=0.05) of launch zone spacing differences found that the 2m instances were significantly lower than the 4m and 5m instances.  

Increasing launch zone spacing to 5m for Building set B resulted in lower median total block durations overall that slightly decreased with increased number of waves, as shown in Figure~\ref{fig:jblm_total_dur_b}. Overall, the smaller the spacing, the longer the median total block durations. The 5 (number of waves) $\times$ 4 (launch zone spacing) between-groups ANOVA related to total block durations identified significant main effects for number of waves ($F(4,12) = 38.34, p < 0.01$), and launch zone spacing ($F(3,12) = 56.61, p < 0.01$). A posthoc Tukey test assessing differences by number of waves found that the 1 and 2 wave results were significantly higher than the 2, 3, 4, and 6 wave results, and the 1 wave results were significantly higher than 2 waves. A posthoc Tukey test of the launch zone spacing indicated that the 2m and 3m spacing had significantly higher block durations than 4m and 5m. Additionally, the 4m spacing block duration was significantly higher than 5m.

Overall, the Leschi Town CACTF's total block duration significantly decreased as the spacing between UAVs increased for both Building sets. Total block duration consistently decreased for Building set A as the spacing increased, irrespective of the number of waves. After 3 waves for Building set B, the effects of increasing the spacing on the total block duration became less prominent. The total block duration for Building set A significantly decreased with 2 waves, but significantly increased with any additional waves. Building set B saw a significant reduction in the total block duration with an increase, up to 3 waves, at which point there was no further reduction in the total block duration. These seemingly counterintuitive results emphasize the importance of the mission plan design's organization of and interdependencies between tactics and their goal locations in reducing congestion.

\afterpage{\clearpage}
\subsection{Pre-FX6: Fort Campbell, Cassidy CACTF} \label{sec:pre_fx_cassidy}
The Pre-FX6 congestion analysis shifted to the Fort Campbell Cassidy CACTF with the finalized FX6 schedule. The evaluation hypotheses remained the same, and the evaluation was conducted in a nearly identical manner to the Leschi Town CACTF evaluation, Section~\ref{sec:pre_fx_jblm}. 

\subsubsection{Experimental Methodology} \label{sec:cassidy_exp_design}

A notable difference for the Cassidy CACTF evaluation was the addition of a new \textbf{independent variable}, \textit{configuration pattern}, which refers to whether the vehicles are placed in the launch zone using a hexagonal or a square configuration. The \textbf{mission plan design} was completed as in Section~\ref{sec:jblm_plan_design}; however, only a single Building set was analyzed, as detailed in Table~\ref{tab:wave_assignments}, due to the added independent variable and limited time before the start of FX6. These two changes primarily affect the launch zone configuration, with minor changes to the evaluation's execution. The \textbf{dependent variables} remained the same, \textit{total block count} and \textit{total block duration}. 

\paragraph{Launch Zone Configuration}
The configuration pattern must be accommodated within the launch zone configuration file. A configuration file was created for each combination of launch zone spacing and configuration pattern (i.e., square and hexagonal), resulting in eight total configuration files. Each square configuration pattern used 6 rows with 10 columns of UAVs, as described in Section~\ref{sec:cassidy_theoretical_analysis}. The analyzed launch zone spacing between vehicles were 2, 3, 4, and 5m. The hexagonal configuration was created  using the square layout, and adjusted the spacing between rows of vehicles to $\textit{Launch zone spacing} \times \sqrt3/2$. Every other vehicle row was shifted by $\textit{Launch zone spacing} / 2$ m laterally (i.e., half a column sideways). These adjustments ensured that vehicles continued to conform to the minimum safety distance requirements, while also consuming less overall space than the square configuration.

\paragraph{Experiment Execution}
Twenty trials were performed for each combination of launch zone spacing, number of waves, and configuration pattern. Five total mission plans were created based on the number of waves. 800 total trials were executed (i.e., 5 mission plans x 4 launch zone spacing values x 2 configuration patterns x 20 trials = 800). Each simulation trial ran for 20 minutes after the final wave was deployed. 

\afterpage{\clearpage}

\subsubsection{Results}

\begin{figure*}[!tbh]
\centering
\begin{subfigure}{0.45\textwidth}
    \centering
    \includegraphics[width=\linewidth]{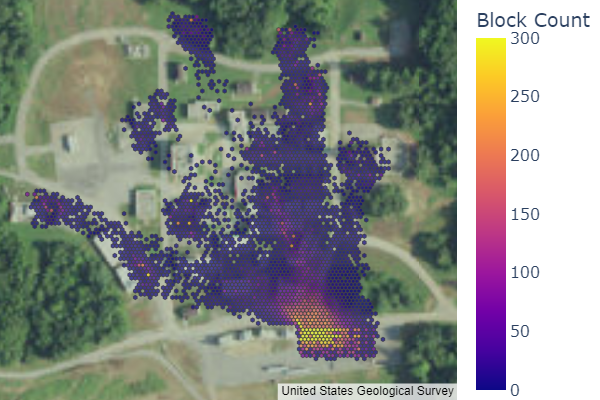}
    \caption{Square - block locations and counts.}
    \label{fig:cassidy_all_square_map}
\end{subfigure}%
\hfill
\begin{subfigure}{0.45\textwidth}
    \centering
    \includegraphics[width=\linewidth]{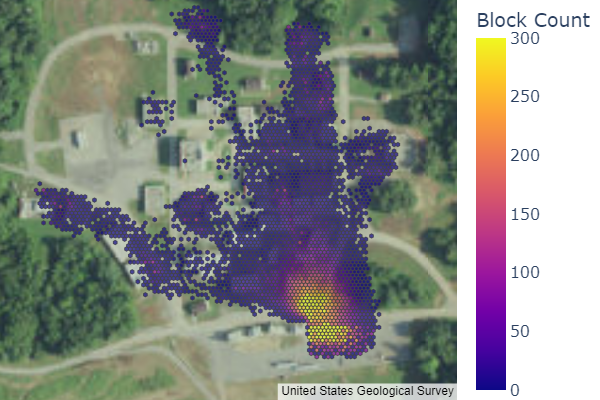}
    \caption{Hexagonal - block locations and counts.}
    \label{fig:cassidy_overall_map}
\end{subfigure}%

\begin{subfigure}{.45\textwidth}
    \centering
    \includegraphics[width=\linewidth]{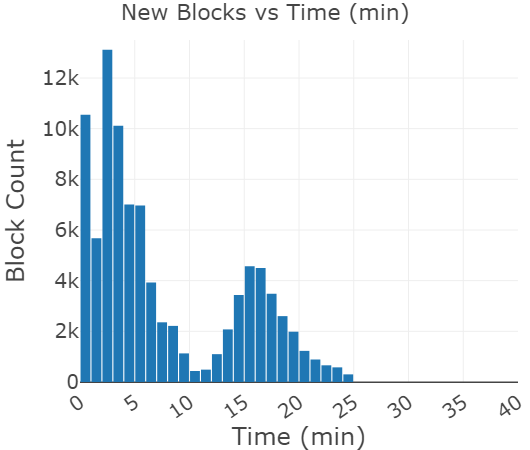}
    \caption{Square - block count by block start time.}
    \label{fig:cassidy_all_square_histogram}
\end{subfigure}
\hfill
\begin{subfigure}{.45\textwidth}
    \centering
    \includegraphics[width=\linewidth]{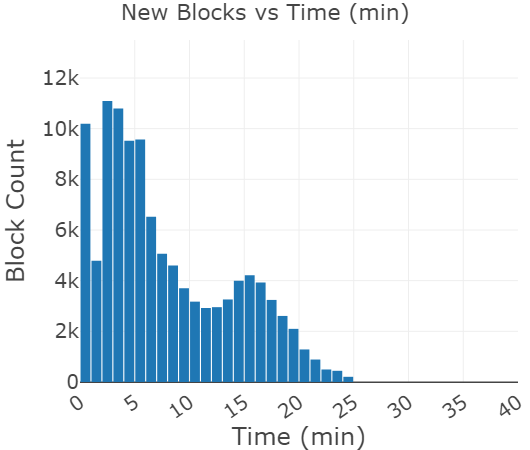}
    \caption{Hexagonal - block count by block start times.}
    \label{fig:cassidy_hex_all_blocks_histogram}
\end{subfigure}

 \caption{The Fort Campbell, Cassidy CACTF simulated congestion total block location heatmaps for the square (a) and hexagonal configuration (b), as well as the total block count by block start time histograms for the square (c) and hexagonal configurations (d).}
\end{figure*}

The Cassidy CACTF heatmaps, by the square, Figure~\ref{fig:cassidy_all_square_map}, and hexagonal configurations, Figure~\ref{fig:cassidy_overall_map}, indicate the locations at which all blockages occurred across all simulation trials, and their frequency. Similar to the Leschi Town CACTF's heatmap, the majority of the congestion occurred in or near the launch area, with congestion also occurring along heavily traveled routes and near the buildings to which Building surveil tactics were assigned. The hexagonal configuration does have a larger number of blocks just north of the launch zone, as compared to the square layout. Generally, the congestion away from the launch area is aligned with the buildings to which Building surveil tactics were assigned. 

Similarly to the Joint Base Lewis McChord Leschi Town CACTF Building set A total blockage start time histogram, the timing of the Cassidy CACTF blockages, shown in Figures~\ref{fig:cassidy_all_square_histogram} and d, occur much more frequently at the start of the mission plans and appear associated with the launch wave timings. The Cassidy CACTF's overall size is about half of the Leschi Town CACTF. Further, the Cassidy CACTF launch zone, as specified pre-FX6, was 520$m^{2}$, or 40\% the size of the Leschi Town's launch zone. The smaller Cassidy launch zone and more compact CACTF led to a substantially larger number of block instances. This larger number of total blockages occurred throughout the mission and across the CACTF. The square configuration generally had more total blockages at the start of the mission, even though there was less congestion at the choke point north of the launch zone. The hexagonal configuration generally had a higher sustained number of new blockages after 7 minutes, which led to this configuration resulting in more total blockages. There is an uptick in blockage instances between 14 and 19 minutes for both configurations, which is associated with UAVs returning to the launch zone due to low battery or task completion. 

\begin{figure*}[!htb]
\centering
\begin{subfigure}{.49\textwidth}
    \centering
    \includegraphics[width=\linewidth]{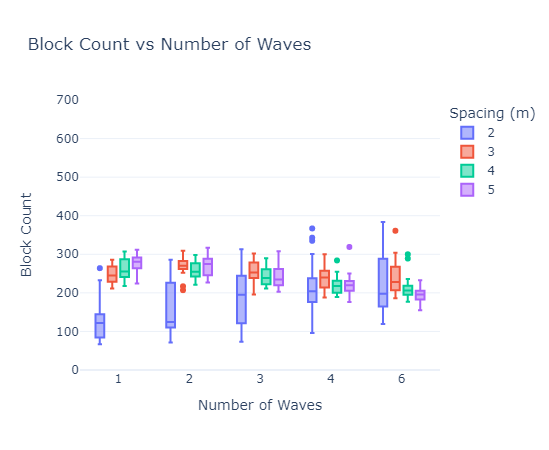}
    \caption{Square layout - block count.}
    \label{fig:cassidy_total_count_a}
\end{subfigure}
\hfill
\begin{subfigure}{.49\textwidth}
    \centering
    \includegraphics[width=\linewidth]{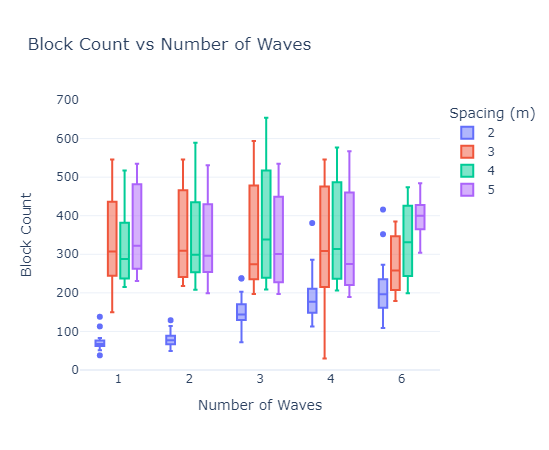}
    \caption{Hexagonal layout - block count.}
    \label{fig:cassidy_total_count_b}
\end{subfigure}

\begin{subfigure}{.49\textwidth}
    \centering
    \includegraphics[width=\linewidth]{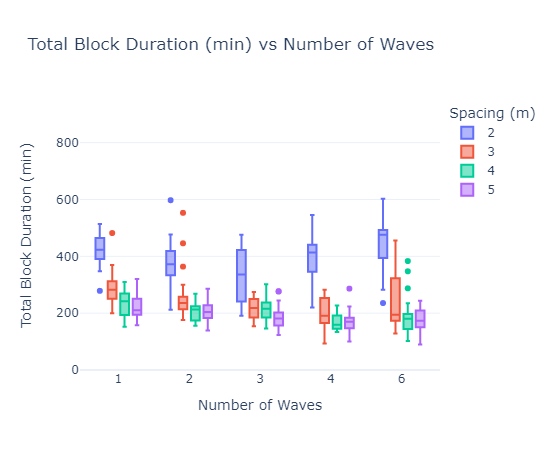}
    \caption{Square layout - block duration.}
    \label{fig:cassidy_total_dur_square}
\end{subfigure}%
\hfill
\begin{subfigure}{.49\textwidth}
    \centering
    \includegraphics[width=\linewidth]{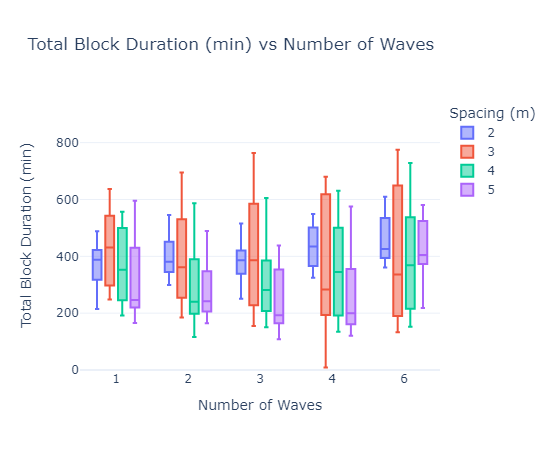}
    \caption{Hexagonal layout - block duration.}
    \label{fig:cassidy_total_dur_hex}
\end{subfigure}%

\caption{Fort Campbell, Cassidy CACTF's Building set congestion results' box plots, median, for block count and duration by layout configuration, number of launch waves and launch zone spacing.}
\end{figure*}

The Cassidy CACTF's median block count results, by configuration pattern, are presented in Figures~\ref{fig:cassidy_total_count_a} and b. Both configurations' results show that a 2m spacing has the lowest block count across all numbers of waves. The differences between the maximum and minimum block counts with the hexagonal configuration was larger for all spacings greater than 2m. A 5 (number of waves) $\times$ 4 (launch zone spacing) $\times$ 2 (configuration pattern) between-groups ANOVA performed for the total block counts yielded significant main effects for launch zone spacing ($F(4,12) = 103.60, p < 0.01$) and configuration pattern ($F(4,12) = 92.38, p < 0.01$). No significant main effects were found for the number of waves. The posthoc Tukey test of the launch zone spacings determined that the total block count for 2m spacing is significantly lower than for 3m, 4m, and 5m spacings.  A posthoc Tukey test of the differences by configuration pattern indicated that the hexagonal configuration had a significantly higher block count than a square configuration. 

The median block duration results by configuration pattern for the Cassidy CACTF are presented in Figures~\ref{fig:cassidy_total_dur_square} and d, respectively.
The square configuration with the 2m and 3m spacings resulted in longer block durations across all numbers of waves, which was also true with the hexagonal layout for 1, 2 and 3 waves. The square configuration overall had less difference between the minimum and maximum block durations.
A 5 (number of waves) $\times$ 4 (launch zone spacing) $\times$ 2 (configuration pattern) between-groups ANOVA performed on the total block durations identified significant main effects for number of waves ($F(4,12) = 10.84, p < 0.01$),  launch zone spacing $F(3,12) = 228.78, p < 0.01$), and configuration pattern ($F(12,12) = 3.18, p < 0.01$). A posthoc Tukey test by the number of waves indicated that 3 waves were significantly lower than the 1 or 6 wave instances. The posthoc Tukey test of launch zone spacings found that 2m instances were significantly higher than the 3, 4, and 5m spacings. Additionally, the total block duration for 3m instances was significantly higher than the 4m or 5m. A posthoc Tukey test of the pairwise differences by configuration pattern discovered a significantly higher block duration for the hexagonal layout. Heatmaps for the Cassidy CACTF results by number of launch waves did not present any substantial differences and are excluded in favor of brevity. 

\begin{figure*}[!htb]
\centering
\begin{subfigure}{.48\textwidth}
    \centering
    \includegraphics[width=\linewidth]{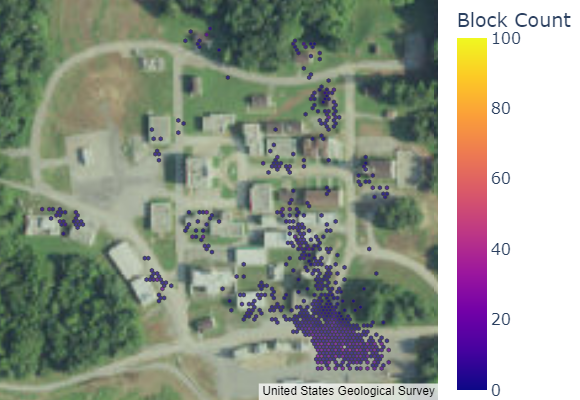}
    \caption{Best-case configuration - block count.}
    \label{fig:cassidy_best_count_map}
\end{subfigure}
\hfill
\begin{subfigure}{.49\textwidth}
    \centering
    \includegraphics[width=\linewidth]{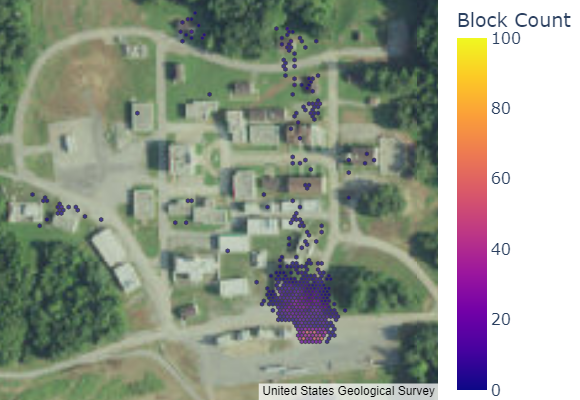}
    \caption{Worst-case configuration - block count.}
    \label{fig:cassidy_worst_count_map}
\end{subfigure}%

\begin{subfigure}{.49\textwidth}
    \centering
    \includegraphics[width=\linewidth]{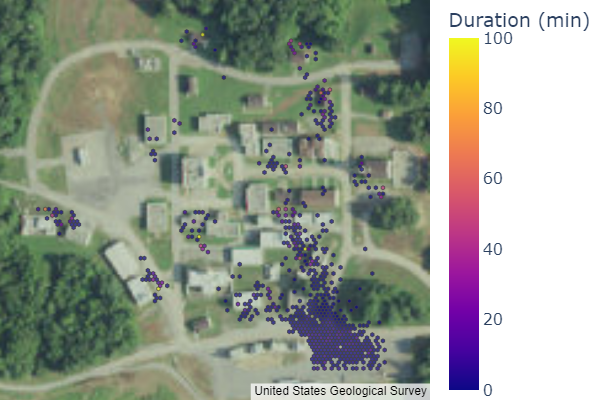}
    \caption{Best-case configuration - block duration.}
    \label{fig:cassidy_best_duration_map}
\end{subfigure}
\hfill
\begin{subfigure}{.49\textwidth}
    \centering
    \includegraphics[width=\linewidth]{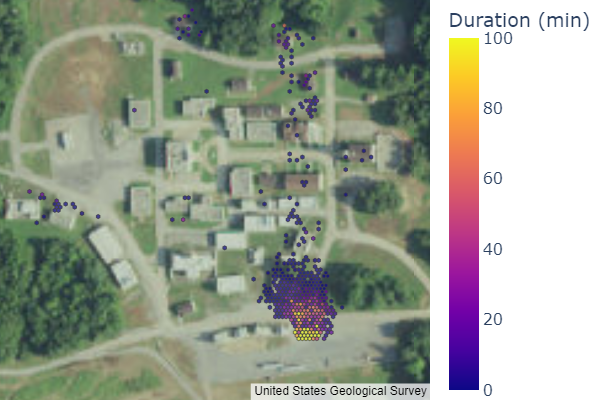}
    \caption{Worst-case configuration - block duration.}
    \label{fig:cassidy_worst_duration_map}
\end{subfigure}%

\caption{The Fort Campbell, Cassidy CACTF's Building set's congestion heatmap for the best-case (square layout, 5m Spacing and 4 waves), and the worst-case (hexagonal layout, 2m spacing and 4 waves) configurations' total block count, (a) and (b) respectively, and  total block duration, (c) and (d) respectively.}
\label{fig:cassidy_best_worst_maps}
\end{figure*}

A comparison of the subjective best-case (square layout, 5m space, 4 waves) and worst-case (hexagonal layout, 2m space, and 4 waves) block counts and block durations are shown in Figure~\ref{fig:cassidy_best_worst_maps}. The results are similar in many ways to the Joint Base Lewis McChord Leschi Town results. The block counts are higher (Figure~\ref{fig:cassidy_worst_count_map}) and have longer durations (Figure~\ref{fig:cassidy_worst_duration_map}) for the worst-case configuration, compared to the same results for the best-case (Figures~\ref{fig:cassidy_best_worst_maps}a and c, respectively). There is a choke point slightly north of the launch zone, which is more prominent with the worst-case configuration. The best-case configuration results in increased congestion at the CACTF's outer regions, which in contrast demonstrates that fewer vehicles in the worst-case configuration navigated beyond the launch zone due to the severe congestion.

\begin{figure*}[!htb]
\centering
\begin{subfigure}{.49\textwidth}
    \centering
    \includegraphics[width=\linewidth]{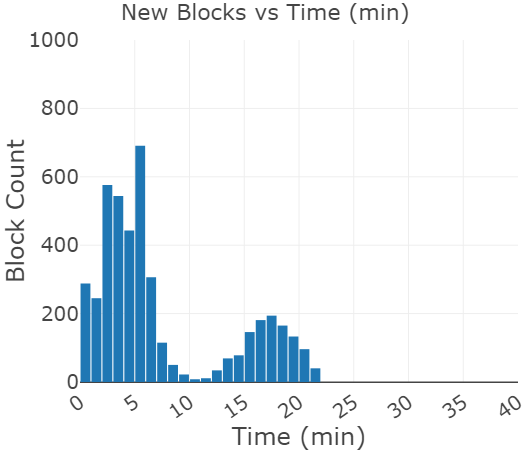}
    \caption{Best-Case configuration - block start time.}
    \label{fig:cassidy_best_histogram}
\end{subfigure}%
\hfill
\begin{subfigure}{.49\textwidth}
    \centering
    \includegraphics[width=\linewidth]{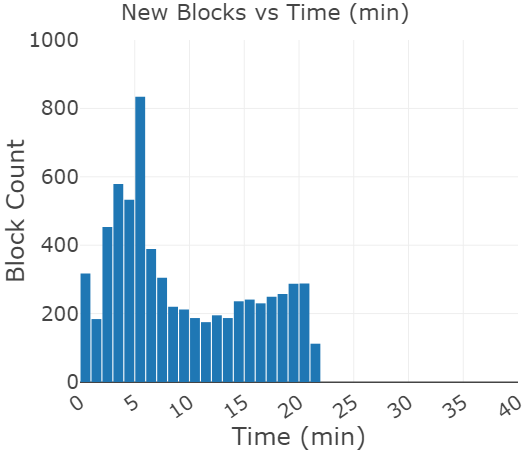}
    \caption{Worst-Case configuration - block start time.}
    \label{fig:cassidy_worst_histogram}
\end{subfigure}%

\caption{The Fort Campbell, Cassidy CACTF's best- (a) and worst-cast (b) block start time histograms. }
\label{fig:cassidy_best_worst_histograms}
\end{figure*}

As with the overall blockage start time histogram, the best- and worst-case blockage start time histograms demonstrate increases in congestion at the start of the mission associated with the timing of each cases' four launch waves, see  Figures~\ref{fig:cassidy_best_histogram} and b, respectively. While the best-case Cassidy configuration shows a steep drop off in new blockages around minute 6, there continues to be new blockages throughout the mission deployment. An increase in new blockages occurs, beginning at the $14^{th}$ minute and continues until the $19^{th}$ minute. This increase is due to UAVs returning to the launch zone after completing their tactics or due to a low battery. The substantially smaller launch zone and the cited choke point both contribute to this increased congestion and new blockage instances. Since both the best- and worst-case instances incorporate four launch waves, the start of the mission for the worst-case is similar in terms of the number of new blockages. While there is an overall decrease in the total blockages after minute 6, the total number of worst-case new blockages is sustained at a higher level throughout the mission as compared to the best-case. 

Overall, the Cassidy CACTF's block count was significantly lower for 2m launch zone spacing than all other spacing values. Additionally, the hexagonal configuration pattern had a significantly higher total block count than the square pattern. The number of waves had no significant impact on the total block count. 

Cassidy's total block duration significantly decreased as the spacing increased. Total block duration for the 4m and 5m spacings were both significantly lower than 2m and 3m. The total block duration was significantly reduced when increasing the number of waves from 1 to 3, but increased with more waves. The total block duration for a hexagonal configuration was significantly higher overall than with the square layout. 

\subsection{Joint Pre-FX Discussion}

The total number of blocks for both the Leschi Town and Cassidy CACTFs increased as the spacing between vehicles increased. This result is counterintuitive until it is compared with the total block duration results. The highest total block duration occurs when the total block count is its lowest, which suggests that as congestion worsens, the number of individual blockages decreases, while their severity may increase significantly. Alternatively, as congestion decreases, the number of blocks may actually increase, but the resulting blockages may not be as severe and may have shorter durations. The total block count alone can be an unreliable congestion metric, for that reason, the total block duration appears to be more reliable for measuring congestion. This phenomenon is visible in Figures~\ref{fig:jblm_best_worst_maps} and \ref{fig:cassidy_best_worst_maps}, where the block duration heatmaps more prominently present the congestion differences between the configurations.

Choke points can arise based on the launch zone area and the CACTF layout, as seen in the heatmaps for both CACTFs. The Leschi Town launch zone is longer and more centrally located with tactics assigned across the breadth of the CACTF, as a result, the launch zone itself becomes a choke point. The same heatmaps for the Cassidy CACTF identify a choke point with the major increase in congestion just north of the launch zone, typically associated with the direction of the buildings to be surveilled. The more compact Cassidy CACTF and launch zone result in this choke point congestion. A larger, in other words longer, launch zone was expected to alleviate this particular choke point. 

\textit{Hypothesis \RomanNumeralCaps{1}} stated that congestion decreases with an increased launch zone spacing, which was supported for both CACTFs. A decrease in congestion occurred for both CACTFs as the launch zone spacing increased from 2m to 4m. Increasing the launch zone spacing further to 5m had little to no effect, except for the Leschi Town's Building set A were 5m spacing further decreased congestion. 

\textit{Hypothesis \RomanNumeralCaps{2}} stated that congestion decreases with more deployment waves, which was partially supported for both the Leschi Town and Cassidy CACTFs. The Leschi Town Building set A's mission plans' congestion improved with 2 and 3 waves, but any additional waves increased congestion 
due to longer duration blockages, 
as visible with the heatmaps in Figure~\ref{fig:jblm_wave_differences}. Leschi Town's Building set B's mission plans' congestion decreased, or was unchanged, with increased waves. Congestion for the Cassidy CACTF mission plans' for the square configuration fell as the number of waves increased for the larger spaces (i.e., $>2$). Further, the maximum and minimum congestion metrics were tighter with the square configuration. The hexagonal configuration's overall values for more than 2m spacing were similar across the spacings and numbers of waves, even though the maximum and minimum values covered a broader range.  

Leschi Town's Building sets provided differing results. No major discrepancies existed between the selected buildings, with similar building location distributions and wave assignment distributions. Although minor differences exist (e.g., buildings 33 and 37 in Building set A, which are on opposite sides of the street at the western end of the launch zone), such selection choices appear to lead to choke points that increase congestion. Two waves balanced the UAV launch zone traffic significantly better than the other wave counts for Building set A, but two waves did not similarly affect the results for Building set B. These changes in congestion with differing numbers of deployment waves are visible in Figure~\ref{fig:jblm_wave_differences}, with a major reduction between 3 waves and either 1 or 6 waves.  Building Set B's congestion reductions occurred with up to 4 waves, with 5 waves performing similarly.

The results support the notion that more waves can reduce congestion; however, too many waves can increase congestion. Blockages tended to begin either at the beginning of the mission or about twenty minutes into the mission (i.e., when vehicles take off and when they return from low battery or tactics completion). Both the Leschi Town and Cassidy CACTFs' results suggest the possibility of identifying an ``optimal'' number of waves for a set of tactic targets; although there is no guarantee that the optimal value will remain the same for different tactic combinations. For example, 2 waves was optimal for Leschi Town Building set A, but 3 waves were optimal for Cassidy's hexagonal configuration. 

\textit{Hypothesis \RomanNumeralCaps{3}} stated that a hexagonal layout will use less overall space than a square layout, while not increasing congestion. This hypothesis was rejected. Even though the hexagonal layout succeeded in using less overall space than the square layout, while maintaining the minimum distance between vehicles, the level of congestion was significantly higher. While not explicitly a requirement of the CCAST system, the additional space provided by a square layout was shown to significantly decrease congestion. The downside of a hexagonal layout can be mitigated by increasing the launch zone spacing.

\afterpage{\clearpage}
\section{Post-FX6 Congestion Analysis}\label{sec:post_fx_analysis}

The FX6 CCAST swarm deployments presented an opportunity to mine the log files to understand the prevalence of congestion by vehicle blockages and blockage durations. The FX mission plans were leveraged to conduct a simulation-based congestion analysis. The evaluation
\textit{Hypothesis \RomanNumeralCaps{4}} states that a simulated and real deployment with identical swarm composition and mission plans will have similar congestion patterns. 

The early, short FX6 integration and dry run shifts focused on validating system capabilities that often incorporate fewer vehicles, minimal mission plans, and minimal simultaneous tactic instantiations. The longer later shifts are intended to deploy the swarm to achieve the mission objectives, but some of those shifts also focus on OFFSET Sprinter integration validations. The sprinters' projects are designed to develop technology the integrator teams potentially need, but are unable to develop themselves. The UAV focused integrated technologies implied that the CCAST's main swarm UAVs were grounded during the sprinter integration testing.  High sustained winds, with wind gusts up to 29 MPH on November 17\textsuperscript{th}, the last shift CCAST deployed as the only swarm in the CACTF, created unique hardware-based challenges that resulted in abnormal mission operations. The remaining FX6 shifts were ``joint shifts'' during which both OFFSET integrator teams deployed their swarms simultaneously on the CACTF. The teams shared their swarm vehicles' telemetry, which populated the shift log files with information that was difficult to differentiate from the CCAST swarm vehicles. Unfortunately, this joint deployment was not announced prior to the FX, so the log files were not adjusted to facilitate the necessary analysis distinctions. 

The November 16\textsuperscript{th} FX6 shift results are analyzed due to a few key characteristics. The number of vehicles staged in the launch zone was 91, 81 UAVs and 10 UGVs. During the shift, 74 unique vehicles were deployed, many of them multiple times. During this shift, the CCAST LTE network did not experience any outages, which resulted in consistent logging to support this analysis. This particular shift's mission plan launched multiple simultaneous Building surveil tactics at the very start of the shift purposely intended to launch as many vehicles as possible, creating a higher likelihood of generating congestion.  
 
The \textbf{launch zone configuration} was created using field notes of the vehicles' staged positions in the FX6 launch zone. The evaluation's \textbf{independent variable} is simply whether the trial was real or simulated. The \textbf{dependent variables} are counts (FX6 data) and descriptive statistics (simulation data) of the timestamped blockages and assigned tactics, as well as the block durations.

\paragraph{Mission Plan Design}
The FX6 mission plan for each shift deployment was logged. The CCAST simulator can execute the FX6 mission plans; however, prior to using the mission plans for the simulated evaluation, a few modifications were necessary. These modifications were made to ensure simulation trials were as true-to-real as possible, and no modifications fundamentally changed the underlying mission plan.

The CCAST system supports live-virtual deployments, in which the swarm has both hardware and virtual vehicles that are treated identically from a system operation perspective. This feature enables increasing the swarm size significantly and facilitates substituting virtual vehicles when conditions (e.g., high winds) constrain the hardware vehicle deployments. This feature provides many benefits, but the virtual UAVs deployed during live-virtual shifts do not encounter congestion or create congestion for the hardware UAVs. The FX6 mission plans often incorporated virtual vehicles. The mission plan's virtual vehicle tactics were independent of the hardware vehicle tactics, meaning no tactics were assigned a mix of virtual and hardware vehicles. This distinction allowed for removing all tactics assigned to the virtual UAVs and UGVs from the FX6's Nov. $16^{th}$ mission plan for use in the simulation evaluation. 
The FX mission plan incorporates hardware UGVs, but they were also excluded from the presented Nov $16^{th}$'s UAV congestion analysis results and were not included in the simulation analysis mission plan.

The mission plan file is saved after the plan is instantiated during an FX shift, which means the dispatcher has allocated hardware vehicles to the plan's tactics. As a result, the saved mission plan incorporates the hardware vehicles' unique identifiers (e.g., tail numbers). These identifier assignments were removed to prevent the dispatcher from attempting to deploy non-existent hardware UAVs during the simulation. 

The CCAST simulator emulates the different swarm UAVs, including their payloads, but the vehicle dynamics are the same regardless of the UAV type. Therefore, only 3DR solos were used as proxies for the hardware UAVs in the launch zone configuration file. 

\paragraph{Experiment Execution}
The November 16\textsuperscript{th} shift's  mission plan supported a two hour deployment. 
The mission plan was designed to deploy as many hardware-only vehicles for as many tactic sorties as possible during this first thirty minutes of the shift; thus, presenting the highest likelihood of generating swarm congestion. The FX6 DARPA provided scenario had a very dense adversarial distribution, which resulted in large numbers of vehicles becoming neutralized quickly. The neutralized UAVs automatically return to the launch zone.  During the shift 74 unique vehicles were deployed, and many were deployed multiple times. A mobile medic was used to revive the neutralized UAVs just before the 15 minute point in the mission. After the UAVs were revived, the swarm commander redeployed them. Therefore, the decision was made to limit the simulation trial runs to a total of sixty minutes, and the overall analysis to the first sixty minutes of the FX6 shift. Twenty trials were performed using the CCAST simulator and the adjusted mission plan. 

\subsection{Results}

\begin{figure*}[!thb]
\centering
\begin{subfigure}{.48\textwidth}
    \centering
    \includegraphics[width=\linewidth]{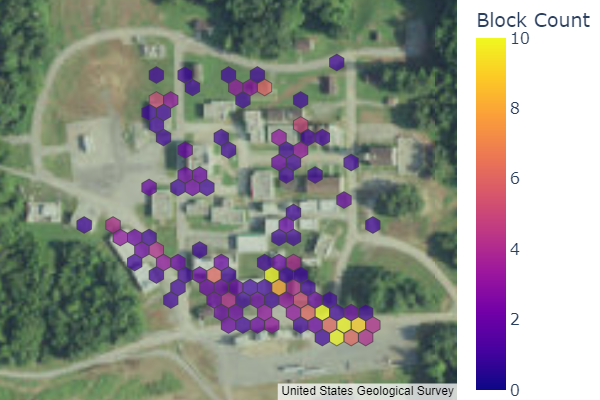}
    \caption{FX6 - total block count.}
    \label{fig:real_count_map}
\end{subfigure}
\hfill
\begin{subfigure}{.49\textwidth}
    \centering
    \includegraphics[width=\linewidth]{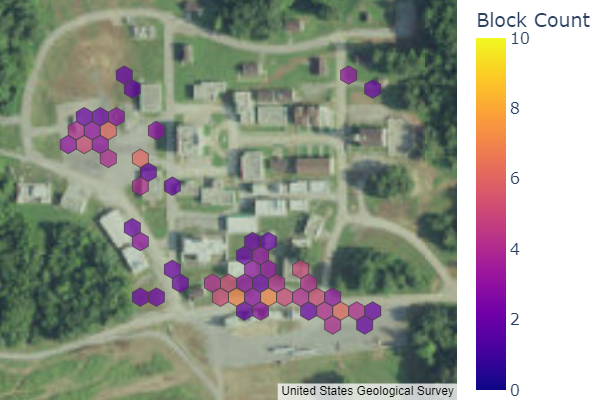}
    \caption{Simulated - total block count.}
    \label{fig:sim_count_map}
\end{subfigure}

\begin{subfigure}{.48\textwidth}
    \centering
    \includegraphics[width=\linewidth]{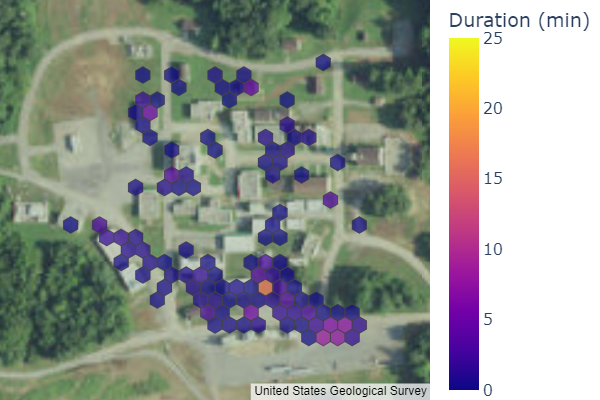}
    \caption{FX6 - total block duration.}
    \label{fig:real_duration_map}
\end{subfigure}
\hfill
\begin{subfigure}{.49\textwidth}
    \centering
    \includegraphics[width=\linewidth]{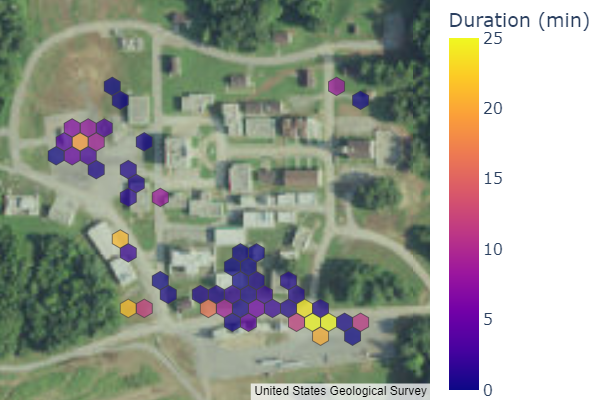}
    \caption{Simulated - total block duration.}
    \label{fig:sim_duration_map}
\end{subfigure}

\caption{Heatmaps of the real FX6 Nov $16^{th}$ shift's block locations (a) and total block durations (c) vs. the simulated shift's block locations (b) and total block durations (d).}
\label{fig:real_v_sim_maps}
\end{figure*}

Heatmaps of the total block count and total block durations for both the actual FX6 Nov $16^{th}$ shift and the simulation evaluation using the shift's mission plan are provided in Figure~\ref{fig:real_v_sim_maps}\footnote{Please note that since there are many fewer block count map instances for this data, the instances are enlarged to make them visible.}. The mission plan deployed as many vehicles as possible at the shift's start; thus, one expects launch zone congestion, which is visible for both the real FX6, see Figure~\ref{fig:real_count_map}, and simulated block counts, see Figure~\ref{fig:sim_count_map}. The initial mission plan for UAVs focused on Building surveil tactics to the left side of the Cassidy CACTF. Blocks occurred across a broader range of CACTF locations during the actual shift due to the swarm commander issuing tactics to vehicles. The block duration heatmaps show some of the longest blockages occurring in the launch area for both the real and simulated results, as shown in Figures~\ref{fig:real_duration_map} and d, respectively. Differences in blockages between the FX6 and the simulation results do exist. For example, no blockages occur in the actual shift results at the building visible in the upper left corner of the figure, but a large number of blockages occurred for that same building with the simulation results. This difference arises from the fact that this building was heavily fortified and the real UAVs were neutralized when near the building, which mitigated the congestion around that building. The simulator does not contain the adversarial artifacts; thus, the virtual UAVs were not neutralized during when near this same building.  

\begin{figure}[hbtp]
\centering
\begin{subfigure}{.47\textwidth}
    \centering
    \includegraphics[width=\linewidth]{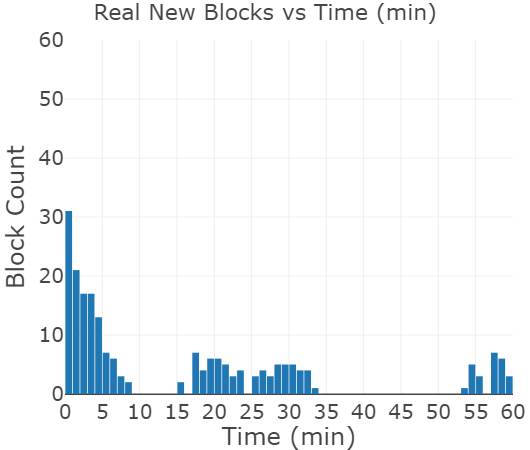}
    \caption{FX6 - total block count.}
    \label{fig:real_new_blocks_v_time}
\end{subfigure}
\hfill
\begin{subfigure}{.47\textwidth}
    \centering
    \includegraphics[width=\linewidth]{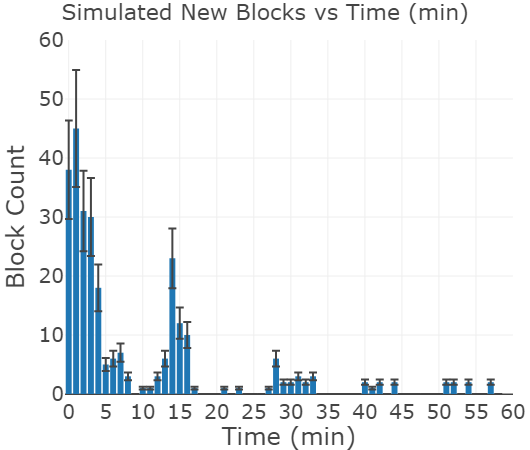}
    \caption{Simulated - mean total block count.}
    \label{fig:sim_new_blocks_v_time}
\end{subfigure}%

\begin{subfigure}{.47\textwidth}
    \centering
    \includegraphics[width=\linewidth]{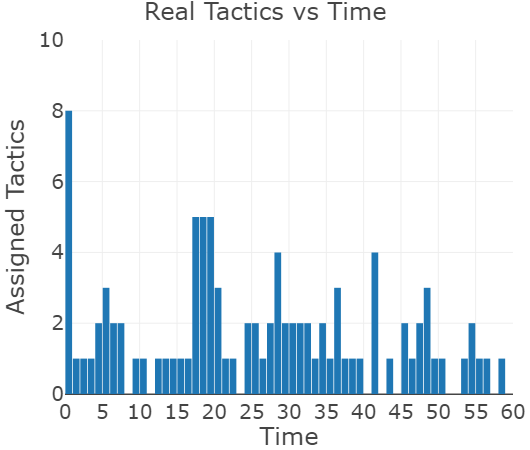}
    \caption{FX6 - total tactic calls.}
    \label{fig:real_tactics_v_time}
\end{subfigure}%
\hfill
\begin{subfigure}{.47\textwidth}
    \centering
    \includegraphics[width=\linewidth]{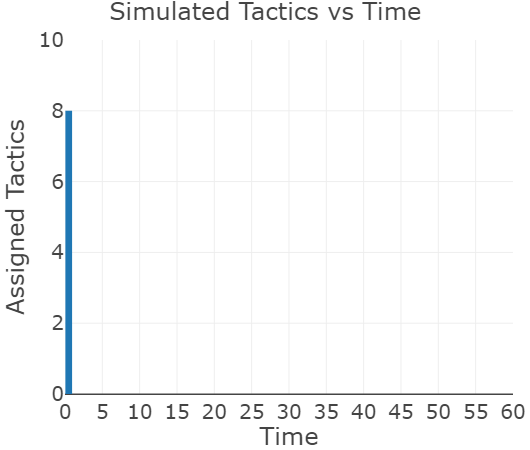}
    \caption{Simulated - mean total tactic calls.}
    \label{fig:sim_tactics_v_time}
\end{subfigure}%

\begin{subfigure}{.47\textwidth}
    \centering
    \includegraphics[width=\linewidth]{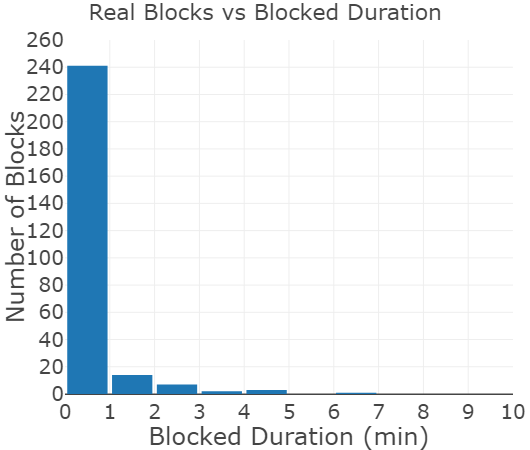}
    \caption{FX6 - total block durations.}
    \label{fig:real_block_dur_v_time}
\end{subfigure}%
\hfill
\begin{subfigure}{.47\textwidth}
    \centering
    \includegraphics[width=\linewidth]{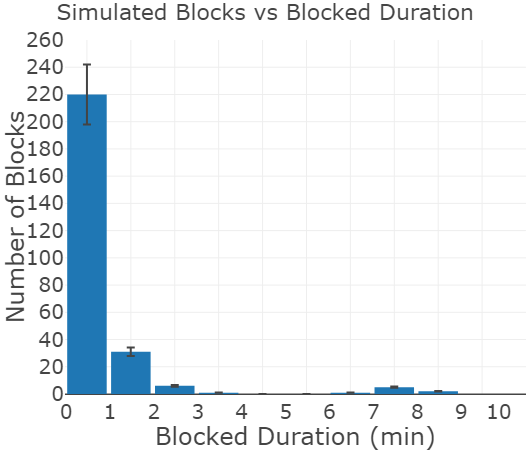}
    \caption{Simulated - mean total block durations.}
    \label{fig:sim_block_dur_v_time}
\end{subfigure}%

\caption{The FX6 Nov $16^{th}$ shift's total block count (a), total tactic calls (c) and total block durations (e) vs.~the simulation's total block count (b), total tactic calls (d), and total block duration (d) means. }
\label{fig:RealvsSim}
\end{figure}

The results histograms facilitate comparisons between the single FX6 Nov $16^{th}$ shift results and the simulation's multiple trials results. The FX6 total block count, assigned tactic count, and total block durations are provided in Figures~\ref{fig:real_new_blocks_v_time}, c, and e, respectively. The corresponding simulation results' across all trials means and standard deviations are provided in Figures~\ref{fig:sim_new_blocks_v_time}, d, and f, respectively.  

The FX6 Phase I mission plan was loaded and deployed at the start of the shift, which is represented in the number of tactics issued between 0 and 1 minute, as shown in Figure \ref{fig:real_tactics_v_time}. While it appears that 8 is a small number of tactics for the mission plan, in fact the eight Building surveil tactics each incorporated the surveillance of multiple buildings. Note, the histogram bucket ranges represent the values greater than or equal to the first number, up to values less than the second value. 
After sending the initial mission plan signal, the swarm commander manually issued a variety of tactics (e.g., Nudging vehicles, Stopping tactics) to relieve the congestion, as well as new tactic assignments for vehicles still in the launch zone or for those vehicles whose tactics had been stopped. These tactics were issued primarily between 5 and 15 minutes, but the number of new UAV blocks was substantially lower during this time frame. The adversary dense scenario resulted in a very large number of UAVs being neutralized during this same time period, which caused neutralized UAVs to RTL. 
The majority of the deployed UAVs had RTL'ed by 15 minutes into the mission, either due to being neutralized, having completed the assigned tactics, or having consumed the available battery power. Hence, the very low number, one, of the new blocks between 10 and 15 minutes.  

At approximately 15 minutes, the mobile medic was used to revive all neutralized UAVs. After which, all UAV batteries were replaced. Once the UAVs are restarted after the battery swap, the swarm commander began issuing tactics to deploy as many vehicles as possible, which is reflected in the number of tactics issued from 15 minutes to 30 minutes, as shown in Figure~\ref{fig:real_tactics_v_time}. The number of blockages increased at the same time, see Figure~\ref{fig:real_new_blocks_v_time}, reflecting the swarm commander's tactic issuing activity. 

The FX6 Phase II mission plan was loaded and deployed 54 minutes into the shift, providing the last opportunity for significant congestion to arise. The FX6 figures reveal that issuing the mission plan's associated tactics generated new blockages. During the remainder of the shift, the swarm commander generated and issued new tactics. While it is known that the mobile medic and another round of battery replacements occurred later in the shift, it was not recorded exactly when those events occurred.

An analysis of the actual FX6 total block durations, see Figure~\ref{fig:real_block_dur_v_time}, indicates that the majority of the durations were short. A total of the 241 block durations were less than or equal to one minute.  The majority (195) of those blocks lasted less than 30 seconds.  Very few blocks had a duration longer than 1 minute. 

The goal of the simulation evaluation was to retain the authenticity of the initial FX6 mission plan, particularly the first 30 minutes; therefore, the swarm commander generated tactics were not included in the simulation evaluation's mission plan. The simulation results show mission plan tactics being issued at the start of the mission plan, as compared to the FX6 mission plan. 
The removal of the large number of swarm commander manually issued tactics leads to the discrepancies between the real and simulated tactics issued.

The simulation's new block count at the start of the mission execution was higher (mean = 38, standard deviation = 9.37), see Figure~\ref{fig:sim_new_blocks_v_time}, than the count of blocks during the FX6 shift, 31. The simulation additionally had a higher mean count of new blocks through the first five minutes.
Since only the mission plan was used for the simulation analysis, no new tactics were created after the mission plan signals were generated at the mission start, as shown in Figure~\ref{fig:sim_tactics_v_time}. Unlike the hardware vehicles, the virtual vehicles were unable to be neutralized; thus, no congestion was generated in the simulation results from this factor. The new blockages between 5 and 20 minutes are most likely due to UAVs RTL'ing, but may also be due to UAVs that were delayed from taking off when the tactic is received. The FX6 and simulated trials block durations were very similar, see Figures~\ref{fig:real_block_dur_v_time} and f, respectively. The FX6 shift results show slightly longer blockage durations, about 20 seconds, as compared to the mean simulation blockage duration. 

The block durations generated by the simulation trials, see Figure~\ref{fig:sim_block_dur_v_time}, were similar to the actual FX6 results in Figure~\ref{fig:real_block_dur_v_time}. 
The most blocks lasted less than one minute. The majority of these block durations were less than or equal to 30 seconds (mean = 203, standard deviation = 23.6). Similar to the FX6 results, the remainder of the block durations that were less than or equal to a minute in duration was substantially smaller (mean = 17, standard deviation = 3.2). Very few blocks had a duration longer than 1 minute. 

Pearson correlations were used to analyze the relationships between the number of tactic calls and the number of generated blocks. A comparison of the FX6 tactic call results with the generated blocks found a positive correlation ($r(59) = 0.4549, p < 0.01$). A similar positive correlation was found for the simulated mission tactic call results compared to the generated block counts ($r(59) = 0.4573, p < 0.01$). 

The Pearson correlations comparing the FX6 results with the simulation trial results for the tactic calls resulted in a positive correlation ($r(59) = 0.5633, p < 0.01$). The analysis of the total block counts from the FX6 results compared to the simulation results found a significant, highly positive correlation ($r(59) = 0.8159, p < 0.01$). Lastly, the analysis of the individual durations of blocks from FX6 compared to the simulation results found a significant, highly positive correlation ($r(19) = 0.9962, p < 0.01$). 

\subsection{Post-FX Discussion}

The results show that \textit{Hypothesis \RomanNumeralCaps{4}} was partially supported across the FX6 and simulation generated results. 
The analyzed mission plans were similar; however, the presence of swarm commander dispatched tactics in the real data lowered the tactics call correlation. The swarm commander generated tactics appeared to have reduced the positive relationship; even so, the real FX6 and simulated blockages showed a strong correlation of when blockages occurred and how long those blockages lasted.

The positive correlation between the data sets supports using simulated analyses to inform pre-deployment decisions that impact mission planning and seek to reduce the impacts of congestion, as in Section~\ref{sec:pre_fx}. This post-FX analysis, with the simulated comparison, focused on the locations, durations, as well as how many and when blocks occurred. Prior to a mission deployment, it will not be possible to know exactly how the adversary will impact the mission plan; therefore, the lack of adversarial components that neutralize the UAVs in the simulation trials is acceptable. The time, cost, and effort associated with deploying a large hardware swarm is and will continue to be very high; thus, increasing the outcomes associated with such deployments and the effectiveness of mission plans is essential. A simulator that emulates the adversarial agents and neutralizes the UAVs was beyond the scope of this program, but is necessary if the intention is to fully understand the potential mission plan's impact on the associated congestion and mission outcomes.  

\afterpage{\clearpage}
\section{Conclusion}

The DARPA OFFSET program requirements that sought to maximize the swarm size, while minimizing the available launch zone area, create one set of constraints resulting in increased blockages between vehicles attempting to depart the launch zone. A CCAST procedural decision to require all deployed vehicles to return to the location from which they deployed in the launch zone upon tactic completion, neutralization, or low power supply, particularly when applied to multi-rotor UAVs, did have some impact on increasing the number of vehicle blockages and associated congestion. This early decision supported two objectives. The first objective was the ability to recover all swarm vehicles within the DARPA specified shift breakdown period. The second objective was to ensure that vehicles returned to a location where human CCAST team members were able to physically replace their batteries during shifts that lasted multiple hours. While the CCAST UGVs can continue working towards achieving mission objectives during the longest field exercise shifts of 3.5 hours, the less expensive, commercially available off-the-shelf UAVs quickly consume a single battery's power supply, on the order of 10 to less than 20 minutes. Achieving the DARPA OFFSET mission objectives necessitates the ability to continually redeploy the UAVs after battery replacements. Two additional interrelated constraints are associated with swarm vehicles'  sensing capabilities. The need to minimize the cost of individual vehicles in order to scale the swarm to 250 vehicles implies that the vehicles' sensor payloads and computation processing capabilities cannot support rapid, accurate detection and avoidance maneuvers, especially for UAVs. As such, the CCAST vehicles, when deployed outdoors, rely on GPS to localize themselves and deconflict with other vehicles. However, the vehicles' relatively small size compared to the larger error associated with the GPS signals, particularly when attempting to avoid mid-air collisions between UAVs in and around the launch zone, resulted in the establishment of minimum safety distances between vehicles during launch zone staging. As the swarm size scales up, the question becomes which constraints can be relaxed to maximize safety and perform the mission, while minimizing congestion. 

The congestion analysis clearly found that 240 vehicles were unable to fit inside the DARPA designated launch zone without violating the CCAST defined safety distances between vehicles that were intended to account for GPS error and avoid mid-air vehicle collisions. Since a mission objective is to deploy large numbers of vehicles simultaneously, congestion will occur. Therefore, additional analyses focused on how to safely deploy a swarm of 240 vehicles using waves of deployments, while also reducing CCAST's safety distance requirements and minimizing congestion in the launch zone.

The total block count metric was found to be an insufficient measure of congestion, as it can lead to incorrect interpretations and is unable to differentiate between blockage severity. The total block duration metric was the more meaningful congestion metric, due to its ability to account for different blockages lasting different durations. Longer total block durations in and around the launch zone led to fewer total blocks, as vehicles were unable to resolve the blockages and move throughout the CACTF. 

Congestion decreased as the distance between platforms increased, with diminishing returns after four meters. The use of deployment waves proved to be another avenue for significantly reducing congestion; however, careful consideration of the number of waves relative to the anticipated UAV tactic assignment deployment durations is critical. The use of too many waves actually increased congestion, even with 90 seconds between waves. 
The optimal number of waves is dependent on the exact composition of mission plans; however, even the use of two waves led to a significant reduction in congestion.

The final DARPA OFFSET field exercise presented an opportunity to compare the incidence of blockages and the severity of congestion from an actual swarm deployment with a simulation of that deployment's mission plan, as a means of demonstrating the efficacy of using the CCAST multi-resolution simulator to analyze congestion mitigation strategies. The strong correlation between the actual and simulated swarm deployments supports using the CCAST simulator to investigate congestion mitigation trade-offs.

The immediately actionable takeaway for deploying swarms in constrained launch areas applies to measuring and reducing robot swarm congestion. First, a smaller launch area led to more congestion and choke points from the launch area into the CACTF. Second, congestion is best assessed using a combination of block count combined with the total block duration. Third, using a more spacious launch zone spacing between vehicles (i.e., 4m or 5m) consistently reduced congestion. Fourth, two deployment waves, and sometimes more, always reduced congestion; however, high numbers of waves with shorter durations between wave launches need to be avoided due to potential increases in congestion. 

\subsubsection*{Acknowledgments}
This research was developed with funding from the Defense Advanced Research Projects Agency (DARPA). The authors thank Drs. Shane Clark and David Diller, Kerry Moffitt, and their CCAST team collaborators from Raytheon BBN and SIFT, LLC.  The views, opinions, and findings expressed are those of the authors and are not to be interpreted as representing the official views or policies of the Department of Defense or the U.S. Government. \\

\noindent DISTRIBUTION STATEMENT A: Approved for public release: distribution unlimited.

\bibliographystyle{apalike}
\bibliography{citation}

\end{document}